%% file: main.tex
\documentclass[10pt,twocolumn,letterpaper]{article}

\usepackage[pagenumbers]{cvpr}

\input{preamble}

\definecolor{cvprblue}{rgb}{0.21,0.49,0.74}
\usepackage[pagebackref,breaklinks,colorlinks,allcolors=cvprblue]{hyperref}
\usepackage{duckuments}
\usepackage{multicol}
\usepackage{multirow}
\usepackage{pifont}
\usepackage{amsmath}
\usepackage{amssymb}
\usepackage{algorithm}
\usepackage{algorithmic}

\def\paperID{31054} %
\def\confName{CVPR}
\def\confYear{2026}

\title{\methodname: 3D Affordance Grounding with Generative Reconstruction}

\author{Chunghyun Park$^1$\qquad Seunghyeon Lee$^{1}$\qquad Minsu Cho$^{1,2}$\\[8pt]
$^1$Pohang University of Science and Technology (POSTECH)\qquad $^2$RLWRLD\\[6pt]
{\small\url{https://chrockey.github.io/Affostruction/}}
}

\begin{document}
\maketitle

\input{sections/0_abstract}

\input{sections/1_intro}
\input{sections/2_related}

\input{sections/3_method}

\input{sections/4_experiments}
\input{sections/5_conclusion}

\section*{Acknowledgments}
This work was supported by the IITP grants (RS-2022-II220113: Developing a Sustainable Collaborative Multi-modal Lifelong Learning Framework (50\%), RS-2022-II220290: Visual Intelligence for Space-Time Understanding and Generation based on Multi-layered Visual Common Sense (20\%), RS-2024-00457882: National AI Research Lab Project (25\%), RS-2019-II191906: AI Graduate School Program at POSTECH (5\%)) funded by the Ministry of Science and ICT, Korea. Chunghyun Park was supported in part by the POSTECHIAN Fellowship.

\input{sections/X_suppl}

\clearpage

{
    \small
    \bibliographystyle{ieeenat_fullname}
    \bibliography{main}
}

\end{document}

%% file: preamble.tex
\usepackage{wrapfig}

\newcommand{\methodname}{Affostruction\xspace}

\newcommand{\cmark}{\ding{51}}  %

\DeclareMathOperator*{\argmax}{arg\,max}

%% file: sections/0_abstract.tex
\begin{abstract}
This paper addresses the problem of affordance grounding from RGBD images of an object, which aims to localize surface regions corresponding to a text query that describes an action on the object.
While existing methods predict affordance regions only on visible surfaces, we propose \methodname{}, a generative framework that reconstructs complete object geometry from partial RGBD observations and grounds affordances on the full shape including unobserved regions.
Our approach introduces sparse voxel fusion of multi-view features for constant-complexity generative reconstruction, a flow-based formulation that captures the inherent ambiguity of affordance distributions, and an active view selection strategy guided by predicted affordances.
\methodname{} outperforms existing methods by large margins on challenging benchmarks, achieving 19.1 aIoU on affordance grounding and 32.67 IoU for 3D reconstruction.
\end{abstract}

%% file: sections/1_intro.tex
\section{Introduction}
\label{sec:intro}

Robotic manipulation requires understanding not only what objects a robot observes, but also how it can interact with them.
Such functional properties---affordances~\cite{gibson2014ecological}---are typically predicted from complete 3D shapes~\cite{3daffordancenet,where2act,adaafford}.
In practice, however, robots observe objects through RGBD cameras from limited viewpoints, resulting in partial observations with significant occlusions.
A robot viewing a mug from the front, for instance, needs to reason about the occluded handle for grasping.
Existing open-vocabulary affordance grounding methods~\cite{ovafields,openad,pointrefer,affordancellm} operate directly on partial point clouds, predicting affordances only on visible surfaces.
Meanwhile, multi-view reconstruction methods~\cite{octmae,neuralrecon,dust3r,mvsnet,neus,neat} and 3D generation methods~\cite{trellis,shap-e,zero1to3} address geometry but not functional affordances. Reconstruction methods recover only observed surfaces, while generation methods can complete unseen regions but operate on single RGB images without leveraging the depth available in robotic settings.
Grounding affordances on occluded surfaces thus requires both completing geometry from depth-conditioned observations and predicting functional properties on the reconstructed shape.

\input{figures/teaser}

We introduce \textbf{\methodname{}}, a generative framework that reconstructs complete object geometry from partial RGBD observations and grounds affordances on the full shape including unobserved regions.
Two properties of the robotic setting further shape our design.
First, robotic scenarios naturally provide multi-view RGBD sequences as cameras move or objects are manipulated.
We exploit this through sparse voxel fusion of DINOv2~\cite{dinov2} features that aggregates information from multiple views while maintaining constant computational complexity, enabling generative reconstruction that extrapolates unseen geometry.
Second, affordances are inherently ambiguous---multiple valid regions exist for the same query (\textit{e.g.}, ``grasp'').
We address this through flow-based generation that learns to model distributions over affordance heatmaps rather than relying on discriminative predictions, naturally capturing the multi-modal nature of functional interactions.
Moreover, the predicted affordances on reconstructed geometry can guide active viewpoint selection, prioritizing views that observe functional regions to progressively improve target region reconstruction and grounding.

We build upon TRELLIS~\cite{trellis}, a foundation model for 3D generation trained on large-scale 3D datasets.
TRELLIS provides strong priors for completing 3D structures, but processes only single RGB images without depth or camera extrinsics, and generates visual appearance rather than functional affordances.
Without depth, it cannot resolve structural details where textures are similar but geometry differs; without camera extrinsics, reconstructions are canonicalized and may not align with the actual input orientation.
We extend it with two components: multi-view sparse voxel fusion that aggregates depth-conditioned DINOv2~\cite{dinov2} features from multiple RGBD views with known camera poses into a constant-size representation, and a flow-based affordance module that generates heatmaps conditioned on text queries over the reconstructed sparse voxels.
These extensions enable \methodname{} to ground affordances on complete geometry from multi-view RGBD observations while leveraging the foundation model's geometric priors.

Experiments on the Affogato~\cite{affogato} benchmark show that \methodname{} achieves 32.67 IoU for 3D reconstruction (54.8\% improvement over the state of the art) and 19.1 aIoU for affordance grounding on complete geometry (40.4\% improvement).
When operating from partial RGBD observations, our method reconstructs complete geometry and grounds affordances on the full shape, including occluded regions, from as few as a single view.
We further leverage predicted affordances to guide active viewpoint selection, prioritizing views that maximize visibility of functional regions.
With just one additional view, this strategy achieves 2.0$\times$ faster improvement over sequential baselines, providing clear gains under limited view budgets.

Our contributions are as follows:
\begin{itemize}[leftmargin=*,topsep=2pt,itemsep=1pt]
\item \textbf{Generative multi-view reconstruction}: We propose sparse voxel fusion of depth-conditioned DINOv2 features with camera extrinsics that enables constant-complexity generative reconstruction aligned with the actual object orientation, extrapolating complete geometry from partial RGBD observations.

\item \textbf{Flow-based affordance grounding}: We introduce a flow-based generative model that predicts affordance heatmaps on reconstructed geometry conditioned on natural language queries, capturing the inherent multi-modality of functional interactions.

\item \textbf{Affordance-driven active view selection}: We select next-best views based on affordance predictions to improve functional region coverage, achieving efficient reconstruction and grounding under limited view budgets.
\end{itemize}

%% file: figures/teaser.tex
\begin{figure}[t]
    \centering
    \includegraphics[width=\linewidth]{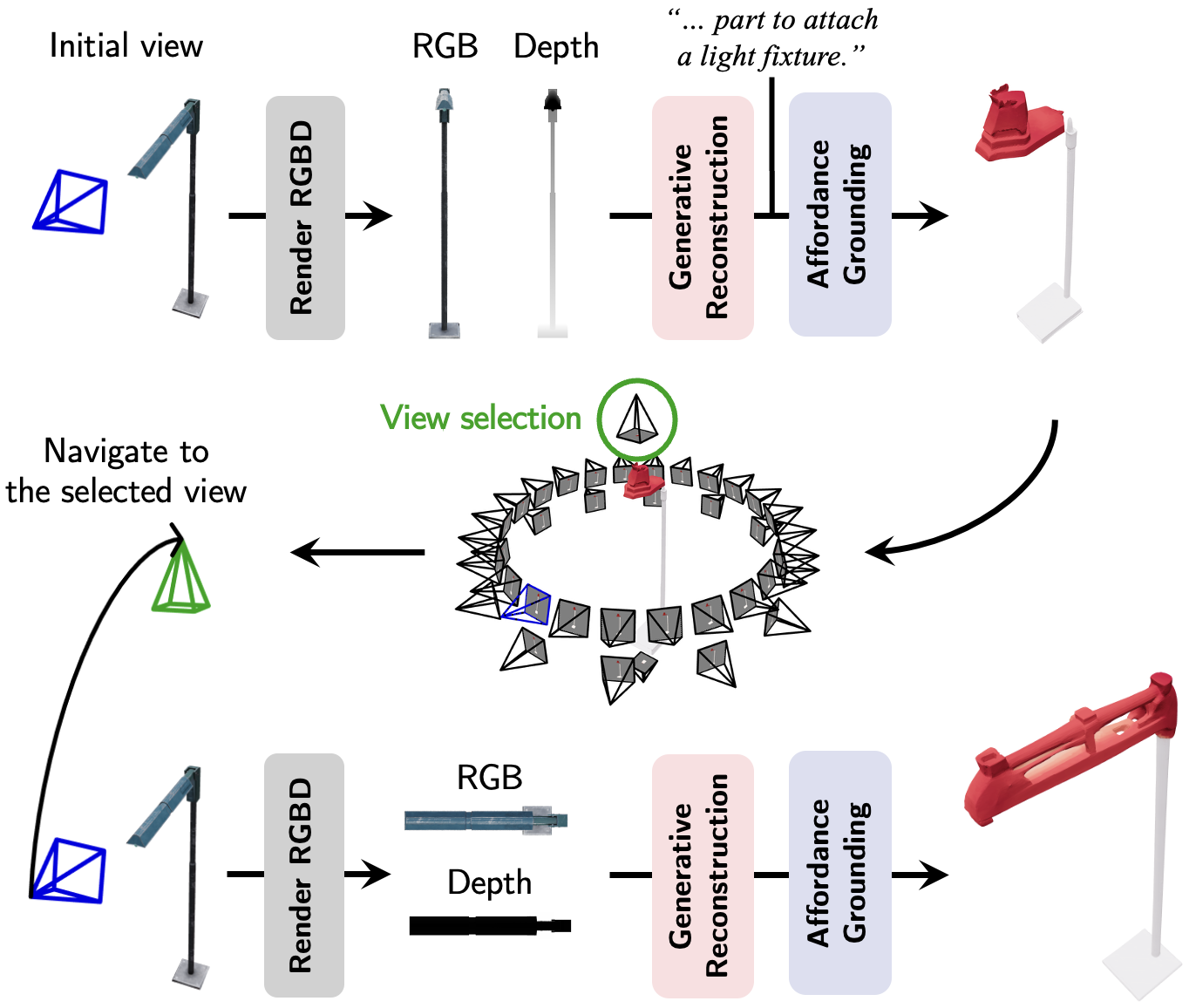}
    \caption{\textbf{\methodname{}.} Given an initial RGBD observation ({\color{blue}blue} camera) where functional regions for an affordance query (\textit{e.g.}, ``attach a light fixture'') are only partially visible or heavily occluded, we reconstruct the complete 3D geometry in a generative manner -- estimating unobserved surfaces -- and ground an affordance region on the full shape effectively. Building on this, an affordance-driven active view selection strategy identifies the most informative next viewpoint ({\color{green!50!black}green} camera). The additional observation acquired from this selected view further refines both the 3D reconstruction and the affordance grounding of the target region.
    }
    \label{fig:teaser}
\end{figure}

%% file: sections/2_related.tex
\input{figures/overview}

\section{Related work}
\label{sec:related}

\noindent\textbf{3D reconstruction.}
Multi-view 3D reconstruction recovers geometry from multiple RGB or RGBD images.
Classical TSDF fusion methods~\cite{bundlefusion,kinectfusion,voxblox,chisel} accumulate depth measurements into volumetric representations, while learning-based approaches such as MVSNet~\cite{mvsnet} and its variants~\cite{transmvsnet,rmvsnet,casmvsnet} predict depth through cost volume matching.
Neural implicit methods~\cite{neus,neuralrecon} learn continuous surface representations from multi-view images but require per-scene optimization, while feed-forward models such as DUSt3R~\cite{dust3r} and MASt3R~\cite{mast3r} predict dense correspondences without camera calibration, enabling direct generalization to unseen scenes.
Despite these advances, all reconstruction methods are inherently limited to observed surfaces and do not extrapolate geometry to unseen regions.
Shape completion methods such as OctMAE~\cite{octmae} and MCC~\cite{mcc} extend reconstruction to unobserved regions through learned priors, but none of these methods predict affordances on the recovered surfaces.

\noindent\textbf{3D generation.}
Image-conditioned 3D generation methods~\cite{shap-e,zero1to3,one-2-3-45,instantmesh,lgm} produce complete shapes from single images via feed-forward prediction or multi-view diffusion.
Sparse voxel generation methods such as TRELLIS~\cite{trellis}, XCube~\cite{xcube}, and 3DTopia-XL~\cite{3dtopia-xl} decompose the problem into structure and appearance stages in 3D latent space.
Multi-view diffusion methods~\cite{dreamcomposer,mvdream,syncdreamer} concatenate per-view tokens, resulting in $O(N)$ token growth that limits inputs to a few views at reduced resolution.
These methods rely solely on RGB inputs without using depth to resolve geometric details where textures are ambiguous, nor do they use camera extrinsics, producing canonicalized outputs that may not align with the actual scene.
None of these methods accept multi-view RGBD inputs or predict functional properties on the generated shapes.

\noindent\textbf{3D affordance grounding.}
3D AffordanceNet~\cite{3daffordancenet} introduced the first 3D affordance grounding benchmark with fixed labels, and interaction-based methods~\cite{where2act,adaafford} learn affordances through simulated contacts on complete shapes.
Open-vocabulary methods~\cite{openad,pointrefer,affordancellm,lmaffordance3d} integrate point cloud features with vision-language embeddings to handle broader queries, and OVA-Fields~\cite{ovafields} extends this to 3D fields.
Affogato~\cite{affogato} scales annotation through automated data generation, and its grounding module Espresso-3D handles diverse open-vocabulary queries.
All these methods assume complete point clouds or predict affordances only on observed surfaces, and their discriminative formulations produce single-mode predictions that cannot capture the multi-modal nature of affordances.
Our method addresses both gaps: it completes geometry from multi-view RGBD observations via sparse voxel fusion with $O(1)$ token complexity, and grounds affordances through flow-based generation that models multi-modal distributions.

%% file: figures/overview.tex
\begin{figure*}[t]
    \centering
    \includegraphics[width=\linewidth]{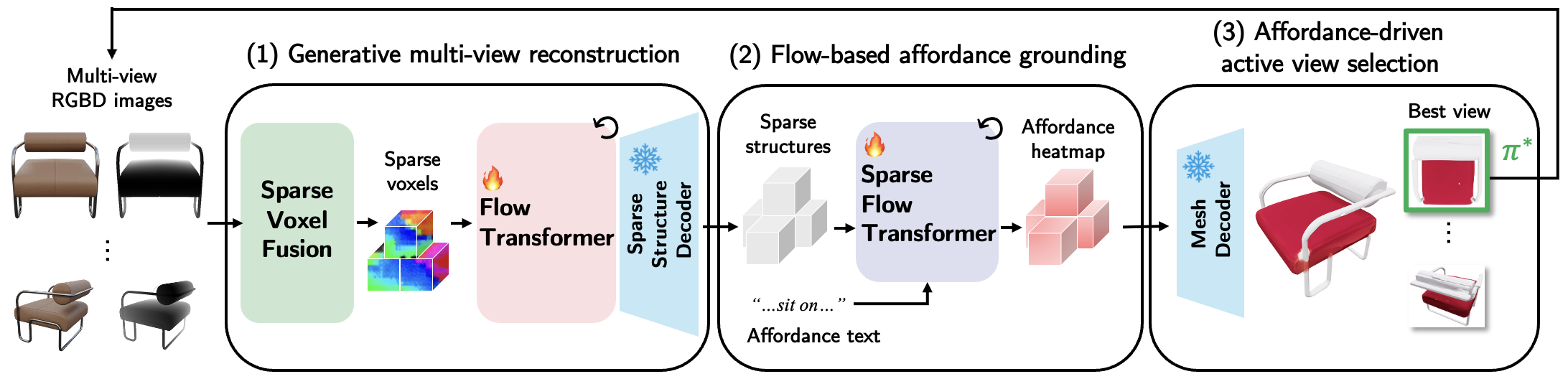}
    \caption{\textbf{\methodname{} overview.} Our approach consists of three stages. (1) Generative multi-view reconstruction: DINOv2~\cite{dinov2} features from RGBD views are fused into sparse voxels using depth and camera parameters, and a flow transformer extrapolates complete structure from partial observations, decoded via a pretrained decoder~\cite{trellis}. (2) Flow-based affordance grounding: a sparse flow transformer conditioned on CLIP~\cite{clip}-encoded text generates affordance heatmaps over reconstructed geometry. (3) Affordance-driven active view selection: next-best viewpoints maximize visibility of high-affordance regions, and a mesh decoder~\cite{trellis} produces the final 3D mesh.}
    \label{fig:overview}
\end{figure*}

%% file: sections/3_method.tex
\section{\methodname{}}
\label{sec:method}

Given multi-view RGBD images with camera parameters, \methodname{} reconstructs complete 3D object geometry---including unobserved regions---and grounds open-vocabulary affordances on it.

As illustrated in Fig.~\ref{fig:overview}, our approach consists of three stages:
(1) Generative multi-view reconstruction (Sec.~\ref{subsec:structure_generation}) fuses features from multiple RGBD views into sparse voxels using depth and camera parameters, extrapolating complete 3D structure from partial observations.
(2) Flow-based affordance grounding (Sec.~\ref{subsec:affordance_grounding}) predicts affordance distributions over the reconstructed geometry conditioned on natural language queries.
(3) Affordance-driven active view selection (Sec.~\ref{subsec:active_sampling}) leverages affordance heatmaps to select viewpoints that prioritize functional regions.

\subsection{Preliminaries: TRELLIS}
\label{subsec:preliminaries}
We build upon TRELLIS~\cite{trellis}, a two-stage rectified flow framework~\cite{rectified_flow} for 3D generation in sparse latent space:

\noindent\textbf{Stage 1: flow transformer} denoises a dense latent tensor $\mathbf{X} \in \mathbb{R}^{r^3 \times C}$ ($r{=}16$) conditioned on DINOv2~\cite{dinov2} patch features $\mathbf{C}$ extracted from a single RGB image. A pretrained sparse structure VAE decoder~\cite{trellis} then converts the denoised tensor into sparse structure $\{\mathbf{p}_i\}_{i=1}^L$, a set of occupied voxel positions where $L \ll r^3$.

\noindent\textbf{Stage 2: sparse flow transformer} denoises a sparse latent tensor initialized at the occupied positions from Stage~1, conditioned on image features, producing structured latent features $\{\mathbf{z}_i\}_{i=1}^L$. A pretrained 3D decoder~\cite{trellis} then renders these into the final 3D representation (\textit{e.g.}, mesh, 3D Gaussians, or radiance fields).

Both stages are trained using the conditional flow matching (CFM) objective~\cite{flow_matching}:
\begin{equation}
\label{eq:flow_matching}
\mathcal{L}_{\text{CFM}}(\theta) = \mathbb{E}_{t, \mathbf{X}_0, \epsilon}\left[||v_\theta(\mathbf{X}_t, \mathbf{C}, t) - (\epsilon - \mathbf{X}_0)||_2^2\right],
\end{equation}
where $\mathbf{X}_t = (1-t)\mathbf{X}_0 + t\epsilon$ interpolates between clean data $\mathbf{X}_0$ and noise $\epsilon \sim \mathcal{N}(0,I)$ with timestep $t \in [0,1]$, and $v_\theta$ predicts the velocity field.

\noindent\textbf{Limitations.}
While effective for single-image 3D generation, TRELLIS has two limitations: (1) it accepts only a single RGB image without depth, limiting accurate structure reconstruction from partial observations; (2) the sparse flow transformer produces visual appearance, not functional affordances. We address (1) by extending the flow transformer to fuse multi-view RGBD inputs (Sec.~\ref{subsec:structure_generation}) and (2) by training a new sparse flow transformer to generate affordance heatmaps from text queries (Sec.~\ref{subsec:affordance_grounding}).

\subsection{Generative multi-view reconstruction}
\label{subsec:structure_generation}
We extend TRELLIS's flow transformer to support multi-view RGBD inputs through \textit{multi-view sparse voxel fusion}. Rather than processing views independently, we aggregate features in 3D using depth guidance before the transformer. This enables generative reconstruction that extrapolates complete geometry from partial observations while maintaining constant token complexity.

\noindent\textbf{Sparse voxel fusion conditioning.}
We represent each view as $\mathcal{V}_i = (I_i, D_i, K_i, T_i)$, where $I_i$ is an RGB image, $D_i$ is a depth image, $K_i$ is camera intrinsics, and $T_i$ is camera extrinsics. For each view, we extract DINOv2~\cite{dinov2} features from $I_i$ and project them to 3D space: each pixel $(u,v)$ with depth $d$ maps to world coordinates $\mathbf{p} = T_i \cdot K_i^{-1}[u, v, d]^\top$, with its DINOv2 feature $\mathbf{f} = \text{DINOv2}(I_i)_{u,v}$. This produces a sparse voxel representation $\mathbf{V}_i = \{(\mathbf{p}_j, \mathbf{f}_j)\}_{j=1}^{M_i}$ for view $i$, where $M_i$ is the number of observed voxels.

To fuse multiple views $\{\mathcal{V}_i\}_{i=1}^N$, we merge their sparse voxel representations. For overlapping voxels at the same position, we average their features; for non-overlapping voxels, we include them in the union. The fused representation $\bar{\mathbf{V}} = \{(\mathbf{p}_m, \bar{\mathbf{f}}_m)\}_{m=1}^{M}$ is then combined with 3D positional encodings to form the final conditioning signal:
\begin{equation}
\mathbf{C}_{\text{voxel}} = \{\bar{\mathbf{f}}_m + \text{PE}_{\text{3D}}(\mathbf{p}_m)\}_{m=1}^{M},
\end{equation}
where $\text{PE}_{\text{3D}}$ is a standard sinusoidal positional encoding extended to 3D voxel coordinates, concatenating separate sinusoidal encodings for each spatial dimension.

\noindent\textbf{Stochastic multi-view training.}
Although TRELLIS's cross-attention conditioning mechanism can handle varying numbers of input tokens at inference, we observe that models trained only on single views exhibit performance degradation when given multiple views (Fig.~\ref{fig:multiview_iou}, left). To address this, we employ stochastic multi-view training: randomly sampling 1-8 views with randomized positions during each training iteration. This enables the model to leverage additional observations, showing consistent improvements as more views are added (Fig.~\ref{fig:multiview_iou}, right).

\noindent\textbf{Training objective.}
We feed the conditioning signal $\mathbf{C}_{\text{voxel}}$ to the flow transformer~\cite{trellis} via cross-attention and train with the same rectified flow objective (Eq.~(\ref{eq:flow_matching})), enabling the model to extrapolate complete 3D structure from partial observations at inference.

\subsection{Flow-based affordance grounding}
\label{subsec:affordance_grounding}

Given the reconstructed 3D geometry and a natural language query (\textit{e.g.}, ``where to grasp''), we predict affordance heatmaps over the complete shape.
Since affordances are multi-modal---a single query admits multiple valid interaction regions---we train a sparse flow transformer from scratch to generate affordance heatmaps as distributions rather than discriminative predictions.

\noindent\textbf{Text conditioning.}
We encode the text query through a pre-trained CLIP~\cite{clip} text encoder to obtain text embeddings:
\begin{equation}
\mathbf{C}_{\text{text}} = \text{CLIP}_{\text{text}}(q),
\end{equation}
where $q$ is the natural language affordance query (\textit{e.g.}, ``where to grasp'') and $\mathbf{C}_{\text{text}} \in \mathbb{R}^{d}$ serves as the conditioning signal.
The affordance flow model denoises a sparse noise tensor $\mathbf{A}_1$ initialized at the voxel positions predicted by Stage 1, conditioned on the text embedding $\mathbf{C}_{\text{text}}$.

\noindent\textbf{Training objective.}
Following the rectified flow formulation, the model predicts velocity fields $v_\theta$ that denoise from noise $\epsilon \sim \mathcal{N}(0,I)$ to clean affordance logits $\mathbf{A}_0 \in \mathbb{R}^{L}$ at $L$ occupied voxels, where the noisy state $\mathbf{A}_t = (1-t)\mathbf{A}_0 + t\epsilon$ with $t \in [0,1]$.
Unlike structure generation which uses MSE loss (Eq.~(\ref{eq:flow_matching})), affordance grounding requires capturing binary occupancy patterns of functional regions.
We therefore define a binary mask loss that combines binary cross-entropy and Dice loss~\cite{dice}:
\begin{equation}
\mathcal{L}_{\text{mask}}(\mathbf{A}', \mathbf{A}) = \mathcal{L}_{\text{BCE}}(\mathbf{A}', \mathbf{A}) + \mathcal{L}_{\text{Dice}}(\mathbf{A}', \mathbf{A}),
\end{equation}
where $\mathbf{A}', \mathbf{A} \in \mathbb{R}^{L}$ are predicted and ground truth affordance logits, respectively (sigmoid transformations are omitted for brevity but applied within BCE and Dice).
This provides both point-wise supervision (BCE) and region-level optimization (Dice).
The flow matching objective with binary mask loss is:
\begin{equation}
\mathcal{L}_{\text{CFM}}(\theta) = \mathbb{E}_{t, \mathbf{A}_0, \epsilon}\left[\mathcal{L}_{\text{mask}}(\epsilon - v_\theta(\mathbf{A}_t, \mathbf{C}_{\text{text}}, t), \mathbf{A}_0)\right].
\end{equation}
Only the affordance flow model is optimized while the structure generation model and text encoder remain frozen.
At inference, we denoise sampled noise conditioned on text query to generate affordance logits, then apply sigmoid to obtain probability heatmaps.

\subsection{Affordance-driven active view selection}
\label{subsec:active_sampling}

Since \methodname{} predicts affordances on reconstructed geometry, the predicted heatmaps can guide subsequent view selection.
Given current observations, we select the next-best viewpoint that maximizes visibility of high-affordance regions, prioritizing functional areas during multi-view capture.
\input{tables/reconstruction_toys4k}

\noindent\textbf{Candidate view generation.}
We generate a set of $K=40$ candidate camera poses $\Pi = \{\pi_1, \ldots, \pi_{K}\}$ uniformly distributed on a hemisphere around the target object.

\noindent\textbf{Affordance visibility metric.}
Given the sparse structure with affordance heatmap $\{(\mathbf{p}_i, a_i)\}_{i=1}^L$ from Sec.~\ref{subsec:affordance_grounding}, we decode it into an affordance-colored mesh $\mathcal{M}$ using the pretrained TRELLIS~\cite{trellis} mesh decoder.
For each candidate pose $\pi_i \in \Pi$, we render this affordance-colored mesh $\mathcal{M}$ to obtain a 2D image $A_{\text{render}}$.
The affordance visibility score is simply the sum of heatmap values in the rendered image:
\begin{equation}
\mathcal{S}(\pi_i, \mathcal{M}) = \sum_{u,v} A_{\text{render}}(u,v),
\end{equation}
where $A_{\text{render}}(u,v)$ is the affordance heatmap value at pixel $(u,v)$ in the rendered image from pose $\pi_i$.
This metric prioritizes viewpoints that observe high-affordance regions.

\noindent\textbf{Iterative view selection.}
We select the next viewpoint by maximizing the visibility score:
\begin{equation}
\pi^* = \argmax_{\pi_i \in \Pi} \mathcal{S}(\pi_i, \mathcal{M}).
\end{equation}
Starting from an initial view $\mathcal{V}_1$, we select the next-best pose $\pi^*$, capture RGBD observation $\mathcal{V}^*$ from $\pi^*$, and add it to the view set.
This process repeats iteratively, with each round of reconstruction and affordance prediction informing the next viewpoint selection.

%% file: tables/reconstruction_toys4k.tex
\begin{table*}[!t]
    \centering
    \caption{\textbf{3D reconstruction results on Toky4K~\cite{toys4k}.} We compare RGB-to-3D generation models and MCC~\cite{mcc}, an RGBD-to-3D reconstruction model. Since MCC does not produce mesh outputs, rendering-based metrics (PSNR, LPIPS) are not available.}
    \label{tab:reconstruction_toys4k}
    \resizebox{0.8\linewidth}{!}{
        \begin{tabular}{lcccccccc}
            \toprule
            \multirow{2}{*}[-2pt]{Method} & \multirow{2}{*}[-2pt]{Depth} & \multicolumn{5}{c}{Geometry} & \multicolumn{2}{c}{Appearance} \\
            \cmidrule(lr){3-7} \cmidrule(lr){8-9}
            & & IoU~$\uparrow$ & CD~$\downarrow$ & F-score~$\uparrow$ & PSNR-N~$\uparrow$ & LPIPS-N~$\downarrow$ & PSNR~$\uparrow$ & LPIPS~$\downarrow$ \\
            \midrule
            Shap-E~\cite{shap-e} &  & ~~6.39 & 0.6724 & 0.0096 & 16.16 & 0.3014 & 13.07 & 0.4358 \\
            InstantMesh~\cite{instantmesh} &  & 13.68 & 0.4063 & 0.0391 & 20.46 & 0.2463 & 16.59 & 0.2989 \\
            LGM~\cite{lgm} & & ~~9.39 & 0.5660 & 0.0267 & 17.03 & 0.3974 & 13.59 & 0.4126 \\
            TRELLIS~\cite{trellis} &  & 19.49 & 0.3694 & 0.0496 & 20.96 & 0.2089 & 17.61 & 0.2435 \\
            \midrule
            MCC~\cite{mcc} & \cmark & 21.11 & 0.3299 & 0.0648 & N/A & N/A & N/A & N/A \\
            \methodname{} (ours) & \cmark & \textbf{32.67} & \textbf{0.2427} & \textbf{0.0997} & \textbf{22.64} & \textbf{0.1421} & \textbf{18.84} & \textbf{0.1922} \\
            \bottomrule
        \end{tabular}
    }
\end{table*}

%% file: sections/4_experiments.tex
\section{Experiments}
\label{sec:experiments}

We evaluate \methodname{} on 3D reconstruction (Sec.~\ref{subsec:reconstruction}), affordance grounding on complete geometry (Sec.~\ref{subsec:complete_affordance}), and affordance grounding from partial observations (Sec.~\ref{subsec:partial_affordance}).
We further analyze multi-view training, active view selection, multi-object scenes, and runtime (Sec.~\ref{subsec:analysis}).

\subsection{Setup}
\label{subsec:exp_setup}

\noindent\textbf{Training datasets.}
For the reconstruction flow transformer, we follow TRELLIS~\cite{trellis} and train on 3D-FUTURE~\cite{3dfuture}, HSSD~\cite{hssd}, ABO~\cite{abo}, and Affogato's train split~\cite{affogato}.
We use Affogato's train split instead of Objaverse-XL~\cite{objaversexl}, which overlaps with Affogato source data, to prevent test set leakage.
The affordance flow transformer is trained on Affogato's train split, which contains affordance annotations.

\noindent\textbf{Evaluation datasets.}
Following TRELLIS~\cite{trellis}, we evaluate 3D reconstruction on 1,250 samples from Toky4K~\cite{toys4k}, a dataset of 4,000 toy objects with ground truth geometry (sampled object IDs provided in supplementary material).
For affordance grounding (both complete and partial settings), we use Affogato's test split~\cite{affogato}.
Partial evaluation uses only the first RGBD view as input.

\noindent\textbf{Implementation details.}
We train for 1M steps on 8 A100 GPUs with batch size 8 per GPU, using AdamW~\cite{adamw} ($\text{lr}=10^{-4}$).
Visual features are extracted using DINOv2-large~\cite{dinov2}, and text queries are encoded with CLIP ViT-L/14~\cite{clip}.
We use classifier-free guidance~\cite{cfg} with 10\% unconditional dropout during training.
Stochastic multi-view training samples 1-8 views per iteration.

\input{tables/affordance_affogato}
\input{tables/haffordance_affogato}
\input{figures/partial_affordance_qual}

\subsection{3D reconstruction}
\label{subsec:reconstruction}
\noindent\textbf{Metrics.}
We measure geometric accuracy via volumetric IoU, Chamfer Distance (CD), and F-score ($\tau=0.05$).
For appearance, we render RGB and normal images from a random view ($r=2$, $\text{FoV}=40^\circ$) and compute PSNR and LPIPS (-N suffix for normals).

\noindent\textbf{Results.}
Table~\ref{tab:reconstruction_toys4k} compares reconstruction quality on Toky4K.
\methodname{} achieves 32.67 IoU, outperforming both TRELLIS~\cite{trellis} (19.49, +67.6\%) and MCC~\cite{mcc} (21.11, +54.8\%).
TRELLIS produces high-fidelity generations but, without depth conditioning, suffers from inconsistencies in object orientation and scale relative to ground truth, resulting in poor geometric metrics.
MCC addresses this through depth-based reconstruction, achieving better geometric accuracy than TRELLIS across IoU, CD, and F-score, though it does not produce mesh outputs for rendering-based evaluation.
Our method surpasses both by combining depth conditioning with flow matching-based generation, showing that generative reconstruction with sparse voxel fusion provides stronger geometric priors than RGB-only generation or discriminative depth-based methods.

\subsection{Complete 3D affordance grounding}
\label{subsec:complete_affordance}

\noindent\textbf{Metrics.}
Following Espresso-3D~\cite{affogato}, we report average Intersection over Union (aIoU), Area Under Curve (AUC), Similarity (SIM), and Mean Absolute Error (MAE).
Among these, we consider aIoU as the primary metric: since affordance regions serve as contact points for robotic manipulation, precise spatial localization matters most.
Accordingly, we train with standard BCE, whereas prior methods~\cite{pointrefer,affogato} adopt focal BCE, which lowers MAE but at the cost of aIoU.

\noindent\textbf{Results.}
Table~\ref{tab:affordance_affogato} evaluates affordance grounding given complete geometry.
We compare against OpenAD~\cite{openad} and PointRefer~\cite{pointrefer} using their official implementations, and Espresso-3D~\cite{affogato} reimplemented following the paper.
Unlike existing methods~\cite{openad,pointrefer,affogato} that fine-tune vision and text encoders with discriminative objectives, our generative approach produces affordance heatmaps without additional encoder training.
While this yields slightly lower AUC (72.0 vs 79.0) and SIM, our method achieves 19.1 aIoU---40.4\% better than Espresso-3D (13.6), indicating more precise localization of affordance regions.
This indicates that generative modeling of affordance distributions achieves stronger spatial grounding than discriminative approaches, even without task-specific encoder fine-tuning.

\subsection{Partial 3D affordance grounding}
\label{subsec:partial_affordance}

\noindent\textbf{Metrics.}
In the partial setting, affordances must be evaluated across the entire 3D geometry including unobserved regions---a requirement for robotic manipulation that needs functional understanding beyond visible surfaces.
While prior work~\cite{openad,pointrefer,affogato} evaluates affordances only within observed regions, we introduce metrics that assess affordance prediction on complete 3D geometry.
We threshold both predicted and ground truth affordance point clouds $\{(\mathbf{p}_i', a_i')\}$ and $\{(\mathbf{p}_j, a_j)\}$ at five probability levels ($\tau \in \{0.1, 0.2, 0.3, 0.4, 0.5\}$) to obtain binary affordance regions, and compute volumetric IoU and Chamfer Distance (CD) between them. We report their averages:
\begin{equation}
\begin{aligned}
\text{aIoU} &= \frac{1}{|\mathcal{T}|} \sum_{\tau \in \mathcal{T}} \text{IoU}(\mathcal{P}_\tau', \mathcal{P}_\tau), \\
\text{aCD} &= \frac{1}{|\mathcal{T}|} \sum_{\tau \in \mathcal{T}} \text{CD}(\mathcal{P}_\tau', \mathcal{P}_\tau),
\end{aligned}
\end{equation}
where $\mathcal{P}_\tau = \{\mathbf{p}_i \mid a_i \geq \tau\}$ are thresholded point sets and $\mathcal{T} = \{0.1, 0.2, 0.3, 0.4, 0.5\}$.
Averaging across multiple thresholds avoids sensitivity to a single binarization criterion.
These metrics jointly evaluate two capabilities: accurate 3D reconstruction and precise affordance localization on the reconstructed geometry---high performance requires both, as accurate reconstruction alone is insufficient if affordance predictions are poorly localized, and affordance predictions cannot compensate for inaccurate geometry.

\noindent\textbf{Results.}
Table~\ref{tab:haffordance_affogato} evaluates affordance prediction on complete objects from single RGBD views.
Existing methods~\cite{openad,pointrefer,affogato} predict affordances only on observed surfaces without reconstruction, yielding 0.38--0.60 aIoU on complete geometry since they cannot account for unobserved regions.
Two-stage pipelines using TRELLIS~\cite{trellis} (RGB) or MCC~\cite{mcc} (RGBD) for reconstruction followed by pretrained affordance models substantially improve over this, with the best reaching 4.74 aIoU (MCC+Espresso-3D), indicating that geometric completion is critical for affordance prediction beyond visible surfaces.
\methodname{} achieves 9.26 aIoU---nearly doubling MCC+Espresso-3D---by performing both tasks in a sparse voxel space.
Figure~\ref{fig:partial_affordance_qual} shows qualitative reconstructed affordances.

\subsection{Analysis}
\label{subsec:analysis}

\input{figures/multiview_iou}

\noindent\textbf{Stochastic multi-view training.}
We compare four conditioning approaches using the same flow transformer: (1) \textit{RGB image patches}: DINOv2 features from RGB only, as in TRELLIS~\cite{trellis}, (2) \textit{RGBD image patches (DINOv2)}: depth maps processed through DINOv2, (3) \textit{RGBD image patches (MCC)}: features from pretrained MCC~\cite{mcc} encoder, and (4) \textit{Sparse voxel fusion (ours)}: multi-view DINOv2 features fused in 3D via depth-guided projection.
Each is trained with either single-view or stochastic multi-view (1--8 views per iteration) supervision.
Figure~\ref{fig:multiview_iou} shows that multi-view training is critical for leveraging additional observations: models trained on single views show minimal improvement or degradation when given multiple inputs at inference, while stochastic multi-view training yields consistent improvement as views increase, plateauing around 6--8 views.
\methodname{} achieves the best results, with IoU improving from 33.32 (single view) to 46.65 (8 views).

\input{figures/view_sampling}

\noindent\textbf{Affordance-driven active view selection.}
We evaluate whether affordance predictions can guide view selection to prioritize functional regions on the Furnitures subset of Affogato's test split, where affordance regions are relatively well-defined.
To simulate a challenging starting condition, we select the initial view with the lowest affordance visibility using ground truth heatmaps.
Given this initial view, we compare three strategies: sequential (predetermined circular trajectory), random (uniform selection), and affordance-driven (ours, selecting views that maximize visibility of predicted high-affordance regions).
Figure~\ref{fig:view_sampling} shows that all methods start from the viewpoint with minimal affordance visibility (${\sim}4.3$ aIoU), and affordance-driven selection achieves the fastest improvement, reaching 9.2 aIoU with one additional view---2.0$\times$ over sequential (4.7) and 1.5$\times$ over random (6.2).
The advantage persists through 4 views (12.4 vs.\ 9.1 sequential and 11.0 random).
By 8 views, active and random sampling converge (13.3 vs.\ 13.2 aIoU), while sequential remains behind at 11.7 aIoU, as its predetermined trajectory may miss functional regions.
Figure~\ref{fig:teaser} illustrates this process on a fireplace queried for placing firewood logs.
From an initial side view where the target region is not visible, the generative prior still produces a reasonable reconstruction and affordance grounding.
This estimate guides selection of a next best view that better observes the affordance region, and incorporating it improves reconstruction and grounding.

\input{figures/multi_object}

\noindent\textbf{Extension to multi-object scenes.}
While our work focuses on object-centric settings, we demonstrate extensibility to multi-object scenes by integrating with SAM3D~\cite{sam3d}.
Given a multi-object scene, SAM3D reconstructs and segments individual objects, and our method then grounds affordances on each object independently.
Figure~\ref{fig:multi_object} shows that \methodname{} successfully grounds affordances on each object in this setting.
Since SAM3D does not support multi-view conditioning, integrating our sparse voxel fusion-based conditioning and active view selection with multi-object reconstruction is an interesting future direction.

\input{tables/efficiency}

\noindent\textbf{Runtime and memory.}
Table~\ref{tab:efficiency} compares runtime and peak memory on a single RTX A6000 GPU.
Our method achieves the fastest runtime (7.20 sec): sparse voxel fusion conditioning reduces sampling to 5 steps (vs.\ 25 for TRELLIS), and joint inference avoids the overhead of separate models.
MCC processes 64$^3$ grid points in 2000-query chunks sequentially, resulting in the slowest runtime (35.37 sec) but the lowest peak memory (4.36 GB).
Overall, \methodname{} achieves the best accuracy (Table~\ref{tab:haffordance_affogato}) with the fastest runtime and comparable memory footprint.

\input{figures/failure_cases}

\noindent\textbf{Failure cases.}
Figure~\ref{fig:failure_cases} illustrates two failure modes.
First, severe occlusion can cause reconstruction errors that propagate to affordance predictions (top); conditioning reconstruction on the affordance text query could provide semantic cues that help recover accurate geometry in such cases.
Second, incorrect initial affordance grounding can mislead active view selection away from the target region (bottom), as subsequent views are chosen based on erroneous estimates; incorporating error detection strategies from semi-supervised learning~\cite{fixmatch,flexmatch} to identify and correct unreliable predictions is a promising future direction.

%% file: tables/affordance_affogato.tex
\begin{table}[!t]
\centering
\caption{\textbf{Complete 3D affordance grounding results on Affogato~\cite{affogato}.} All methods receive ground-truth complete geometry as input. aIoU is the primary metric for spatial localization.}
\label{tab:affordance_affogato}
\resizebox{0.95\linewidth}{!}{
    \begin{tabular}{lcccc}
        \toprule
        Method & aIoU~$\uparrow$ & AUC~$\uparrow$ & SIM~$\uparrow$ & MAE~$\downarrow$ \\
        \midrule
        OpenAD~\cite{openad} & ~~3.1 & 64.8 & 0.329 & 0.150 \\
        PointRefer~\cite{pointrefer} & 10.5 & 76.1 & 0.405 & 0.120 \\
        Espresso-3D~\cite{affogato} & 13.6 & \textbf{79.0} & \textbf{0.429} & \textbf{0.111} \\
        \methodname (ours) & \textbf{19.1} & 72.0 & 0.426 & 0.217 \\
        \bottomrule
    \end{tabular}
}
\end{table}

%% file: tables/haffordance_affogato.tex
\begin{table}[!t]
    \centering
    \caption{\textbf{Partial 3D affordance grounding results on Affogato~\cite{affogato}.} Methods without reconstruction predict affordances only on observed surfaces. Two-stage pipelines pair a reconstruction model with a pretrained affordance model.}
    \label{tab:haffordance_affogato}
    \resizebox{\linewidth}{!}{
        \begin{tabular}{lcrc}
            \toprule
            Method & Recon. & aIoU~$\uparrow$ & aCD~$\downarrow$ \\
            \midrule
            OpenAD~\cite{openad} &  & 0.38 & 0.4165 \\
            PointRefer~\cite{pointrefer} &  & 0.53 & 0.3072 \\
            Espresso-3D~\cite{affogato} &  & 0.60 & 0.2885 \\
            \midrule
            TRELLIS~\cite{trellis} + OpenAD~\cite{openad} & \cmark & 1.49 & 0.1671 \\
            TRELLIS~\cite{trellis} + PointRefer~\cite{pointrefer} & \cmark & 2.05 & 0.1576 \\
            TRELLIS~\cite{trellis} + Espresso-3D~\cite{affogato} & \cmark & 2.23 & 0.1568 \\
            \midrule
            MCC~\cite{mcc} + OpenAD~\cite{openad} & \cmark & 3.34 & 0.1503 \\
            MCC~\cite{mcc} + PointRefer~\cite{pointrefer} & \cmark & 4.19 & 0.1397 \\
            MCC~\cite{mcc} + Espresso-3D~\cite{affogato} & \cmark & 4.74 & 0.1354 \\
            \methodname (ours) & \cmark & \textbf{9.26} & \textbf{0.1044} \\
            \bottomrule
        \end{tabular}
    }
\end{table}

%% file: figures/partial_affordance_qual.tex
\begin{figure*}[t]
    \centering
    \includegraphics[width=0.9\linewidth]{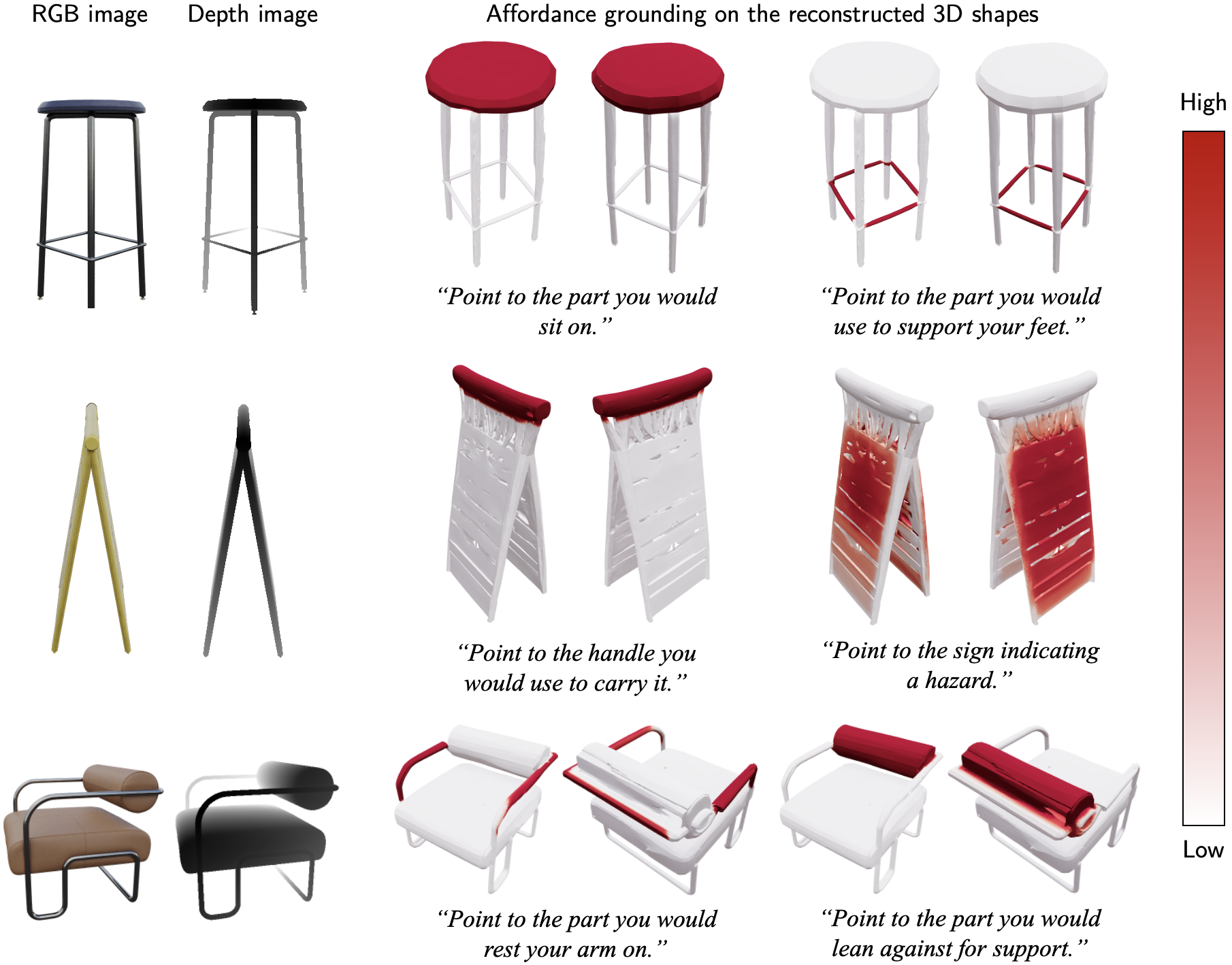}
    \caption{\textbf{Qualitative results on partial 3D affordance grounding.} Affostruction reconstructs complete geometry and grounds affordances throughout entire objects from single RGBD views. Despite limited observations, our method predicts affordances on occluded regions, demonstrating the ability to reason about 3D functional interactions even when large portions of objects are unobserved.}
    \label{fig:partial_affordance_qual}
\end{figure*}

%% file: figures/multiview_iou.tex
\begin{figure}[t]
    \centering
    \includegraphics[width=\linewidth]{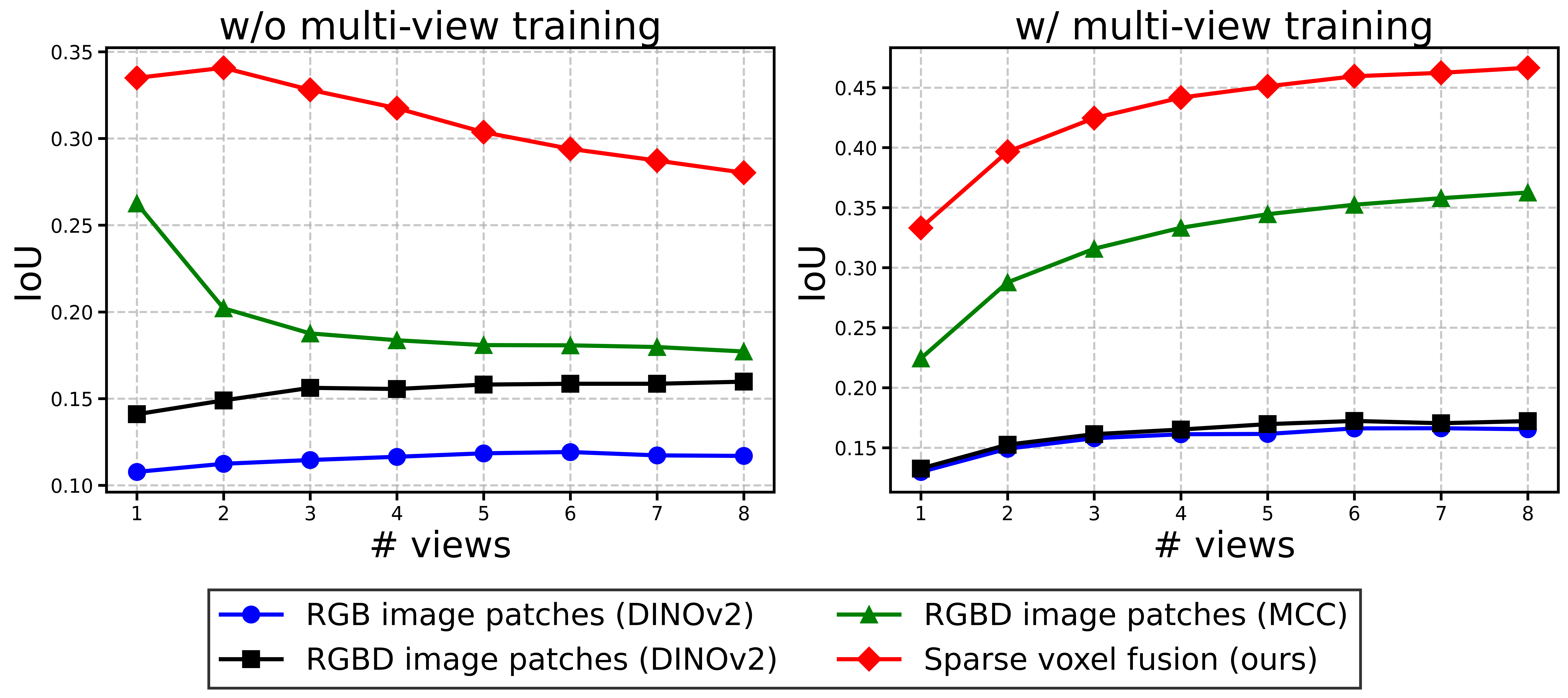}
    \caption{\textbf{Impact of multi-view training.} Reconstruction IoU as a function of input views. Single-view trained models (left) show minimal gains from additional views at inference, while stochastic multi-view training (right) enables consistent improvement. Our sparse voxel fusion achieves the best performance in both settings.}
    \label{fig:multiview_iou}
\end{figure}

%% file: figures/view_sampling.tex
\begin{figure}[t]
    \centering
    \includegraphics[width=0.75\linewidth]{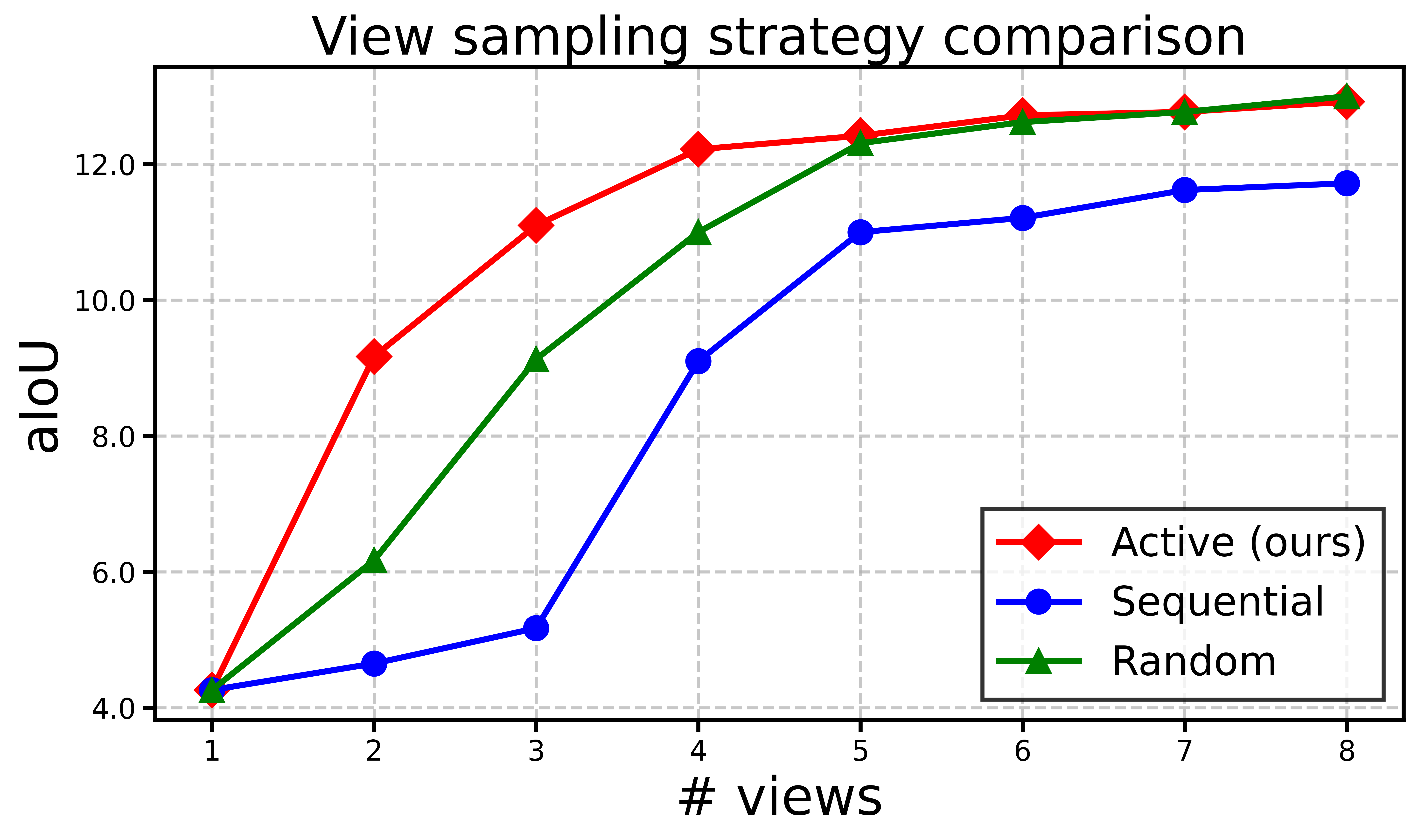}
    \caption{\textbf{Quantitative results of active view selection.} Affordance grounding quality (aIoU) as views are incrementally added from minimal affordance visibility. Affordance-driven selection (red) achieves the fastest improvement, sequential sampling (blue) improves slowest due to its fixed trajectory, and random sampling (green) converges with active selection given more views.}
    \label{fig:view_sampling}
\end{figure}

%% file: figures/multi_object.tex
\begin{figure}[t]
    \centering
    \includegraphics[width=\linewidth]{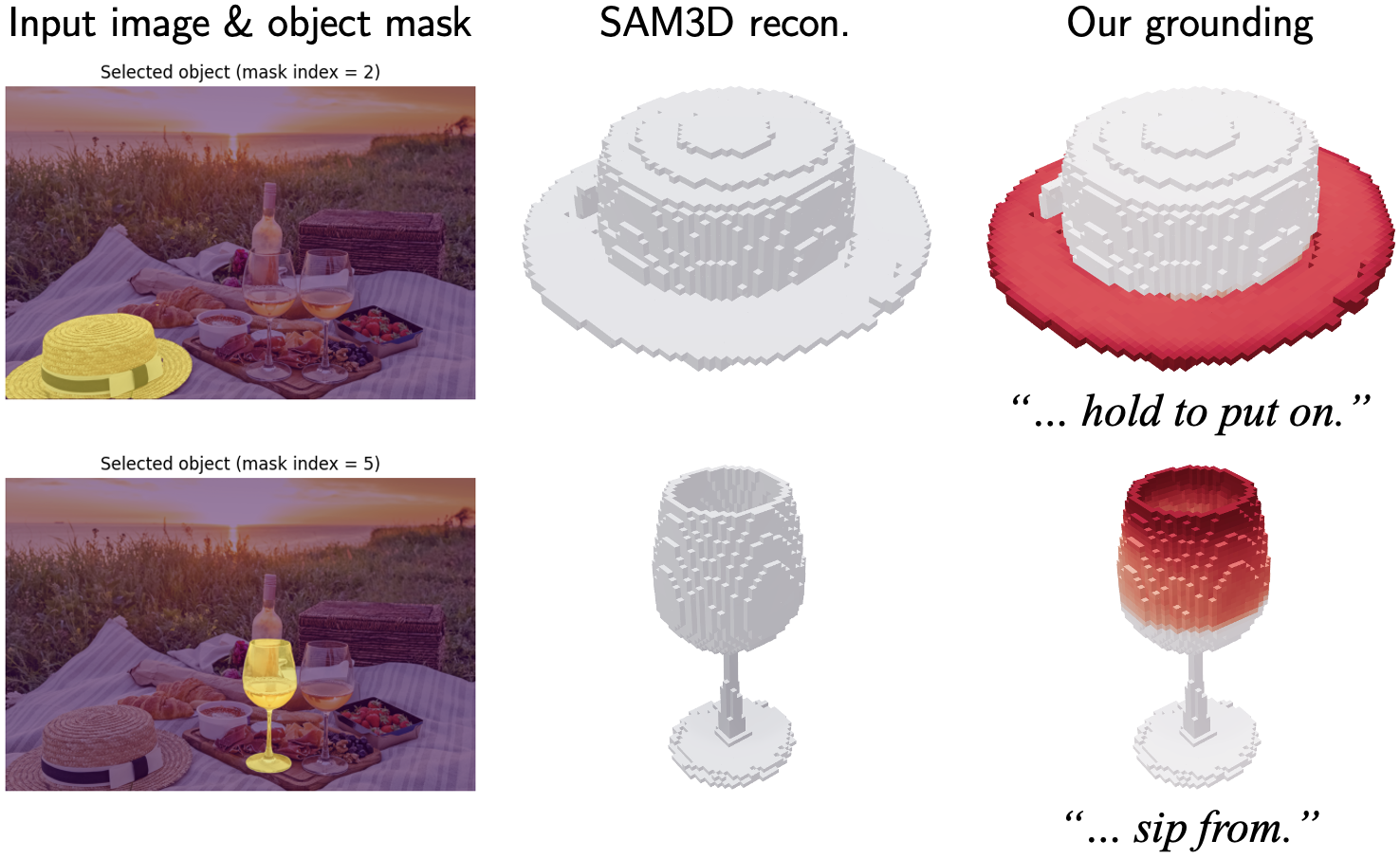}
    \caption{\textbf{Extension to multi-object scenes.} Given a multi-object scene, SAM3D~\cite{sam3d} reconstructs and segments individual objects, and our method grounds affordances on each object independently.}
    \label{fig:multi_object}
\end{figure}

%% file: tables/efficiency.tex
\begin{table}[t]
    \centering
    \caption{\textbf{Runtime and memory comparison.} Average runtime (sec) and peak memory (GB) on Affogato~\cite{affogato}, measured on a single RTX A6000 GPU.}
    \label{tab:efficiency}
    \resizebox{\linewidth}{!}{
        \begin{tabular}{lrrrrrrr}
            \toprule
            \multirow{2.5}{*}{Method} & \multicolumn{3}{c}{Runtime (sec) $\downarrow$} & \multicolumn{3}{c}{Memory (GB) $\downarrow$} \\
            \cmidrule(lr){2-4} \cmidrule(lr){5-7}
            & Rec. & Aff. & Total & Rec. & Aff. & Peak \\
            \midrule
            TRELLIS + Esp.-3D & 7.91 & 0.39 & 8.30 & 16.90 & 4.36 & 16.90 \\
            MCC + Esp.-3D & 34.98 & 0.39 & 35.37 & 1.39 & 4.36 & \textbf{4.36} \\
            \methodname{} (ours) & 4.74 & 2.46 & \textbf{7.20} & 16.35 & 1.26 & 16.35 \\
            \bottomrule
        \end{tabular}
    }
\end{table}

%% file: figures/failure_cases.tex
\begin{figure}[t]
    \centering
    \includegraphics[width=\linewidth]{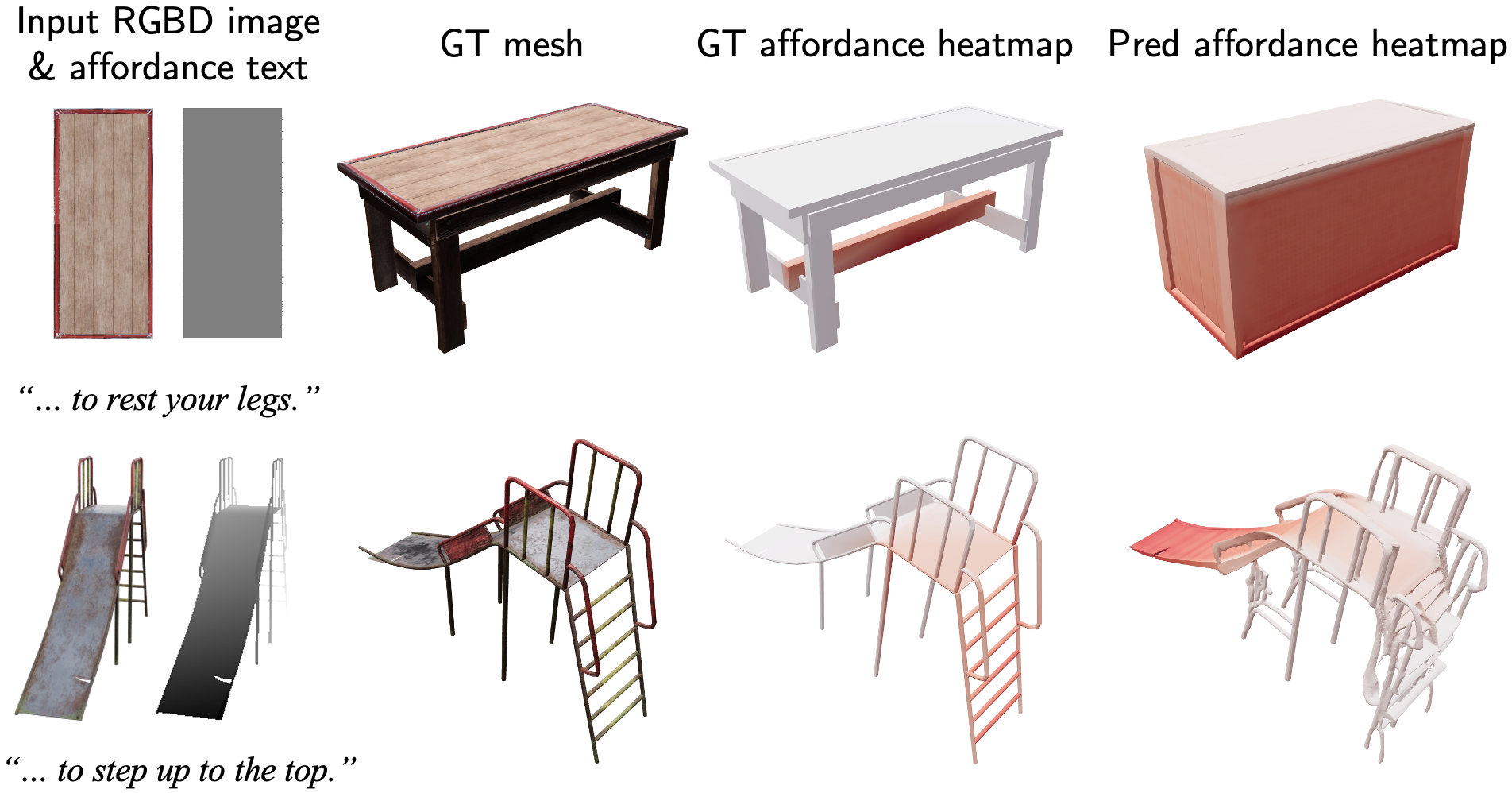}
    \caption{\textbf{Failure cases.} (top) Challenging views with severe occlusion lead to reconstruction errors that propagate to affordance predictions. (bottom) Incorrect initial affordance grounding misleads active view selection away from the target region.}
    \label{fig:failure_cases}
\end{figure}

%% file: sections/5_conclusion.tex
\section{Conclusion}
\label{sec:conclusion}
We have presented \methodname{}, a generative framework that completes object geometry and grounds affordances beyond observed surfaces from partial RGBD observations.
Sparse voxel fusion enables constant-complexity multi-view reconstruction, flow matching captures diverse affordance distributions on the reconstructed geometry, and affordance-driven view selection prioritizes functional regions under limited view budgets.
These components form a closed loop: affordance predictions guide view selection, which in turn improves both reconstruction and grounding.
Experiments demonstrate consistent improvements over existing methods in reconstruction, affordance grounding, and active view selection, with the generative formulation additionally capturing diverse interaction patterns that reflect affordance ambiguity.
Current limitations include dependence on initial affordance estimates for guiding view selection and the object-centric assumption that requires segmentation in multi-object scenes.
Addressing these through error-aware view selection and joint multi-object reconstruction would broaden applicability.
As the method reconstructs functional regions from limited views, applying predicted affordances to robotic manipulation for interaction planning on completed geometry is a natural next step.

%% file: sections/X_suppl.tex
\setcounter{table}{0}
\setcounter{figure}{0}
\setcounter{section}{0}
\renewcommand{\thetable}{A\arabic{table}}
\renewcommand{\thefigure}{A\arabic{figure}}
\renewcommand{\thesection}{\Alph{section}}

\section{Implementation details}

\subsection{Model architectures}

Our method employs two flow-based models: a flow transformer for multi-view 3D reconstruction (Stage 1) and a sparse flow transformer for affordance grounding (Stage 2). Both follow the rectified flow framework~\cite{rectified_flow,trellis} but operate on different representations.

\noindent\textbf{Flow transformer (Stage 1).}
The flow transformer extends TRELLIS~\cite{trellis} to support multi-view RGBD inputs through sparse voxel fusion conditioning. It processes a dense noise tensor $\mathbf{X} \in \mathbb{R}^{16^3 \times 8}$ (4096 tokens) conditioned on fused DINOv2~\cite{dinov2} features from multiple views. We use DINOv2-ViT-L/14 with registers (dinov2\_vitl14\_reg) as the visual feature extractor, producing 1024-dimensional features. Complete architectural specifications are provided in Table~\ref{tab:flow_transformer_arch}.

\noindent\textbf{Sparse flow transformer (Stage 2).}
The sparse flow transformer operates on sparse voxel positions predicted by Stage 1, generating affordance heatmaps conditioned on natural language queries. We use CLIP-ViT-L/14~\cite{clip} (openai/clip-vit-large-patch14) as the text encoder, producing 768-dimensional text embeddings. Unlike the dense reconstruction model, this sparse formulation processes only occupied voxels ($L \ll 4096$), enabling efficient affordance prediction. Full specifications are provided in Table~\ref{tab:sparse_flow_transformer_arch}.

\begin{table*}[t]
\centering
\caption{\textbf{Flow transformer architecture (Stage 1: Multi-view Reconstruction).} The model generates dense 3D structure from multi-view RGBD observations through DINOv2 sparse voxel fusion conditioning.}
\label{tab:flow_transformer_arch}
\begin{tabular}{l|c|l}
\toprule
\textbf{Component} & \textbf{Value} & \textbf{Description} \\
\midrule
\multicolumn{3}{c}{\textit{Transformer Architecture}} \\
\midrule
Resolution & 16 & Spatial resolution of dense 3D grid ($16^3 = 4096$ tokens) \\
Input channels & 8 & Channels of input noise tensor \\
Output channels & 8 & Channels of denoised output tensor \\
Model channels & 768 & Hidden dimension of transformer blocks \\
Conditioning channels & 1024 & Dimension of DINOv2 features (ViT-L/14) \\
Number of blocks & 12 & Depth of DiT (Diffusion Transformer) backbone \\
Number of heads & 12 & Multi-head attention heads per block \\
MLP ratio & 4 & Hidden dimension multiplier for feed-forward layers \\
Patch size & 1 & Spatial patch size for tokenization \\
Positional encoding & APE & Absolute positional encoding \\
QK RMS norm & \checkmark & RMS normalization for query-key projections \\
Precision & FP16 & Mixed precision training with FP16 \\
\midrule
\multicolumn{3}{c}{\textit{Conditioning Mechanism}} \\
\midrule
Visual encoder & DINOv2-ViT-L/14-reg & Feature extractor for RGBD views \\
Feature dimension & 1024 & Output dimension of DINOv2 features \\
Voxel resolution & 16 & Resolution for sparse voxel fusion \\
Image size & $224 \times 224$ & Input image resolution for DINOv2 \\
Max views & 8 & Maximum number of views during training \\
Fusion method & Average & Feature averaging for overlapping voxels \\
\bottomrule
\end{tabular}
\end{table*}

\begin{table*}[t]
\centering
\caption{\textbf{Sparse flow transformer architecture (Stage 2: Affordance Grounding).} The model generates affordance heatmaps on sparse 3D structure conditioned on text queries via CLIP.}
\label{tab:sparse_flow_transformer_arch}
\begin{tabular}{l|c|l}
\toprule
\textbf{Component} & \textbf{Value} & \textbf{Description} \\
\midrule
\multicolumn{3}{c}{\textit{Transformer Architecture}} \\
\midrule
Resolution & 64 & Spatial resolution for latent representation \\
Input channels & 1 & Single channel for affordance heatmap \\
Output channels & 1 & Single channel affordance prediction \\
Model channels & 768 & Hidden dimension of transformer blocks \\
Conditioning channels & 768 & Dimension of CLIP text embeddings (ViT-L/14) \\
Number of blocks & 12 & Depth of DiT backbone \\
Number of heads & 12 & Multi-head attention heads per block \\
MLP ratio & 4 & Hidden dimension multiplier for feed-forward layers \\
Patch size & 2 & Spatial patch size for tokenization \\
I/O residual blocks & 2 & Number of input/output residual blocks \\
I/O block channels & 128 & Hidden channels in I/O residual blocks \\
Positional encoding & APE & Absolute positional encoding \\
QK RMS norm & \checkmark & RMS normalization for query-key projections \\
Precision & FP16 & Mixed precision training with FP16 \\
\midrule
\multicolumn{3}{c}{\textit{Conditioning Mechanism}} \\
\midrule
Text encoder & CLIP-ViT-L/14 & Text feature extractor (openai/clip-vit-large-patch14) \\
Feature dimension & 768 & Output dimension of CLIP text embeddings \\
\bottomrule
\end{tabular}
\end{table*}

\subsection{Training configuration}

\noindent\textbf{Common training setup.}
Both models share core training parameters: 450K steps with batch size 8 per GPU, AdamW optimizer~\cite{adamw} (learning rate $10^{-4}$, no weight decay), EMA (rate 0.9999), and mixed precision (FP16) with adaptive gradient clipping (max norm 1.0, 95th percentile). Timestep $t$ is sampled from a logit-normal distribution ($\mu=1.0, \sigma=1.0$). We apply 10\% unconditional training for classifier-free guidance~\cite{cfg}.

\noindent\textbf{Stage 1: Multi-view reconstruction.}
The reconstruction model is trained with MSE loss and validated on Toky4K~\cite{toys4k} (primary metric: MSE). During training, we randomly sample 1--8 views per object to ensure robust multi-view fusion at inference time.

\noindent\textbf{Stage 2: Affordance grounding.}
The affordance model is trained with a combination of binary cross-entropy and Dice loss, suited for the binary nature of affordance labels. We use elastic training with a linear memory controller (target ratio 0.75) to handle variable structure sizes. The noise scale is set to 5.0 to account for the binary distribution of affordance heatmaps. Validation uses Affogato~\cite{affogato} (primary metric: average IoU).

\subsection{Evaluation configuration}

\noindent\textbf{3D reconstruction on Toky4K~\cite{toys4k}.}
Following TRELLIS~\cite{trellis}, we evaluate reconstruction quality on 1,250 randomly selected samples (SHA256 identifiers in Table~\ref{tab:toys4k_test_samples}). For image metrics, we render both ground truth and predictions from a fixed viewpoint (radius $r=2.0$, FOV $40°$, resolution $512 \times 512$) to compute PSNR and LPIPS. For point cloud metrics (Chamfer Distance and F-score), we render depth maps from 100 views uniformly distributed via Hammersley sampling, unproject to 3D coordinates, and sample 100K points.

\noindent\textbf{Affordance grounding on Affogato~\cite{affogato}.}
We evaluate on the entire test split, following standard protocol: first view and first query per sample. Since our method is generative, we use a reduced noise scale of 0.5 at inference (compared to 5.0 during training) to obtain more consistent predictions for quantitative evaluation.

\begin{table*}[t]
\centering
\caption{\textbf{Toky4K test set samples (SHA256 identifiers).} The 1,250 samples used for 3D reconstruction evaluation, following TRELLIS~\cite{trellis}. Full list available in \texttt{toys4k\_test\_ids.txt}. Hashes truncated to first 12 characters for display.}
\label{tab:toys4k_test_samples}
\footnotesize
\begin{tabular}{@{}lllll@{}}
\toprule
\multicolumn{5}{c}{\textbf{SHA256 Object Identifiers (1,250 samples)}} \\
\midrule
\texttt{\scriptsize 000a283e3a4e...} & \texttt{\scriptsize 002d00832905...} & \texttt{\scriptsize 0036c7bf5fa3...} & \texttt{\scriptsize 00b614f80a13...} & \texttt{\scriptsize 0100555a135f...} \\
\texttt{\scriptsize 016be2974e32...} & \texttt{\scriptsize 019335038b79...} & \texttt{\scriptsize 01a79ca24eac...} & \texttt{\scriptsize 01ac5979fed3...} & \texttt{\scriptsize 02065ccd7123...} \\
\texttt{\scriptsize 021c0a67be93...} & \texttt{\scriptsize 0262655e3219...} & \texttt{\scriptsize 0268f36995da...} & \texttt{\scriptsize 0289dd8d108d...} & \texttt{\scriptsize 0290334c3684...} \\
\texttt{\scriptsize 02a87d37f648...} & \texttt{\scriptsize 02ba6532f9de...} & \texttt{\scriptsize 02c70213d5af...} & \texttt{\scriptsize 02e84388b24c...} & \texttt{\scriptsize 02e9faa6bff3...} \\
\multicolumn{5}{c}{\ldots (1,230 additional samples)} \\
\bottomrule
\end{tabular}
\end{table*}

\section{Sampling parameters search}

\begin{figure*}[t]
    \centering
    \includegraphics[width=0.9\linewidth]{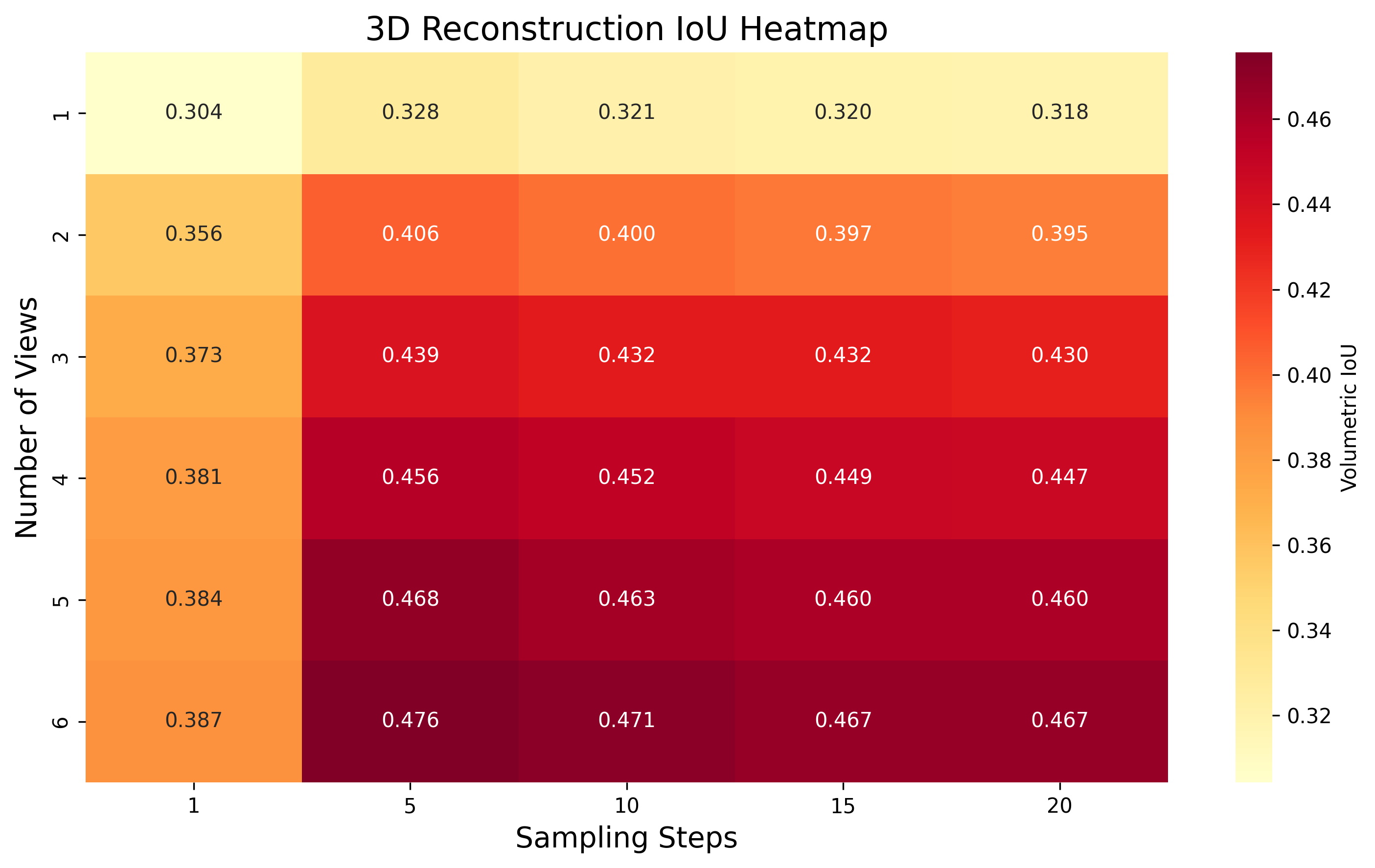}
    \caption{\textbf{Sampling step ablation across different number of views.} We evaluate volumetric IoU for varying sampling steps (1, 5, 10, 15, 20) with 1--6 input views. Reconstruction quality saturates at 5 steps across all view configurations, achieving $5\times$ faster sampling (0.25s) compared to the default 25 steps in TRELLIS~\cite{trellis}.}
    \label{fig:supp_iou_heatmap}
\end{figure*}

We search for the optimal number of sampling steps for the multi-view reconstruction model. Figure~\ref{fig:supp_iou_heatmap} shows volumetric IoU across different sampling steps (1, 5, 10, 15, 20) and number of input views (1--6).
Reconstruction quality plateaus at 5 steps regardless of the number of views, with additional steps providing diminishing returns at increased computational cost. At 5 steps, our model achieves a sampling time of approximately 0.25 seconds, which is $5\times$ faster than the 25-step default (1.29s) of TRELLIS~\cite{trellis}. This efficiency is important for active view selection, where rapid reconstruction enables iterative viewpoint refinement in robotic applications.

\section{Additional qualitative results}

\begin{figure*}[t]
    \centering
    \includegraphics[width=0.9\linewidth]{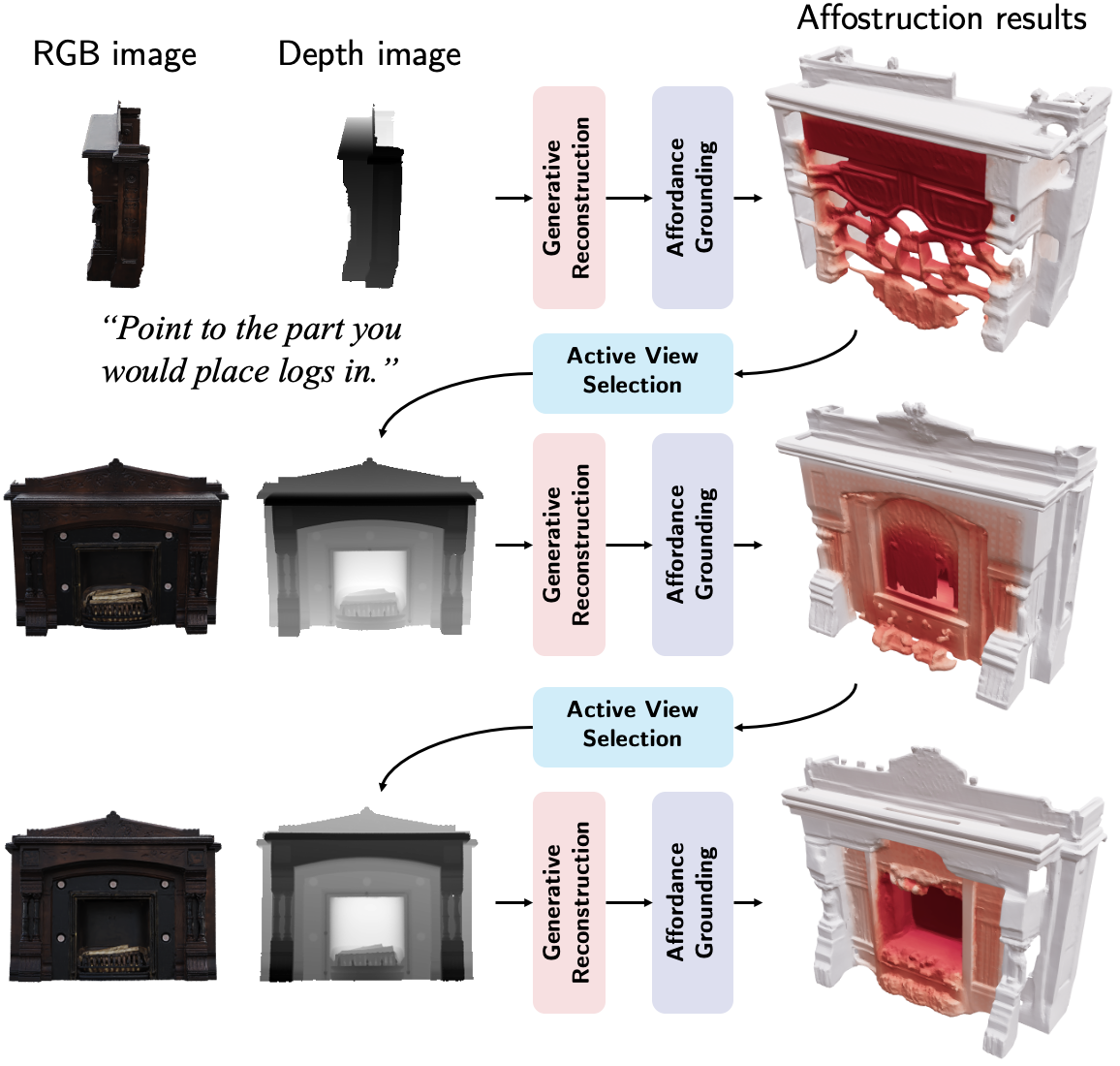}
    \caption{\textbf{Progressive improvement through active view selection.} Starting from challenging viewpoints where target areas are barely visible, \methodname{} progressively improves through iterative steps: (1) generative reconstruction extrapolates complete structure from partial observations, (2) affordance prediction localizes functional regions on the reconstructed geometry, and (3) active view selection targets informative viewpoints based on predicted affordances. As more views are accumulated through multi-view fusion, both reconstruction quality and affordance localization improve. Only the selected view is shown for clarity.}
    \label{fig:supp_afforstruction_qual}
\end{figure*}

Figure~\ref{fig:supp_afforstruction_qual} illustrates the iterative refinement process of \methodname{} starting from challenging initial observations.
We select starting viewpoints where target functional areas have minimal visibility to test the system under difficult conditions.
As views are actively selected based on predicted affordances, the accumulated observations lead to more complete reconstruction, which in turn enables more accurate affordance prediction on the recovered geometry.
Both geometric quality and affordance localization progressively improve as more informative views are incorporated.

\clearpage

%% file: main.bib
@String(IJCV = {Int. J. Comput. Vis.})

@String(CVPR= {IEEE Conf. Comput. Vis. Pattern Recog.})

@String(ICCV= {Int. Conf. Comput. Vis.})

@String(ECCV= {Eur. Conf. Comput. Vis.})

@String(TOG= {ACM Trans. Graph.})

@String(ICLR = {Int. Conf. Learn. Represent.})

@String(CVPRW= {IEEE Conf. Comput. Vis. Pattern Recog. Worksh.})

@String(IJCV  = {IJCV})

@String(CVPR  = {CVPR})

@String(ICCV  = {ICCV})

@String(ECCV  = {ECCV})

@String(TOG   = {ACM TOG})

@String(ICLR  = {ICLR})

@String(CVPRW= {CVPRW})

@inproceedings{kinectfusion,
  title={KinectFusion: Real-time dense surface mapping and tracking},
  author={Newcombe, Richard A and Izadi, Shahram and Hilliges, Otmar and Molyneaux, David and Kim, David and Davison, Andrew J and Kohli, Pushmeet and Shotton, Jamie and Hodges, Steve and Fitzgibbon, Andrew},
  booktitle={IEEE International Symposium on Mixed and Augmented Reality (ISMAR)},
  year={2011}
}

@inproceedings{voxblox,
  title={Voxblox: Incremental 3d euclidean signed distance fields for on-board mav planning},
  author={Oleynikova, Helen and Taylor, Zachary and Fehr, Marius and Siegwart, Roland and Nieto, Juan},
  booktitle={IEEE/RSJ International Conference on Intelligent Robots and Systems (IROS)},
  year={2017}
}

@inproceedings{chisel,
  title={Chisel: Real time large scale 3d reconstruction onboard a mobile device using spatially hashed signed distance fields},
  author={Klingensmith, Matthew and Dryanovski, Ivan and Srinivasa, Siddhartha and Xiao, Jizhong},
  booktitle={Robotics: Science and Systems (RSS)},
  year={2015}
}

@inproceedings{rmvsnet,
  title={Recurrent mvsnet for high-resolution multi-view stereo depth inference},
  author={Yao, Yao and Luo, Zixin and Li, Shiwei and Shen, Tianwei and Fang, Tian and Quan, Long},
  booktitle={IEEE/CVF Conference on Computer Vision and Pattern Recognition (CVPR)},
  year={2019}
}

@inproceedings{casmvsnet,
  title={Cascade cost volume for high-resolution multi-view stereo and stereo matching},
  author={Gu, Xiaodong and Fan, Zhiwen and Zhu, Siyu and Dai, Zuozhuo and Tan, Feitong and Tan, Ping},
  booktitle={IEEE/CVF Conference on Computer Vision and Pattern Recognition (CVPR)},
  year={2020}
}

@inproceedings{neus,
  title={NeuS: Learning neural implicit surfaces by volume rendering for multi-view reconstruction},
  author={Wang, Peng and Liu, Lingjie and Liu, Yuan and Theobalt, Christian and Komura, Taku and Wang, Wenping},
  booktitle={Advances in Neural Information Processing Systems (NeurIPS)},
  year={2021}
}

@inproceedings{dust3r,
  title={DUSt3R: Geometric 3D vision made easy},
  author={Wang, Shuzhe and Leroy, Vincent and Cabon, Yohann and Chidlovskii, Boris and Revaud, Jerome},
  booktitle={IEEE/CVF Conference on Computer Vision and Pattern Recognition (CVPR)},
  year={2024}
}

@inproceedings{mast3r,
  title={Grounding image matching in 3D with MASt3R},
  author={Leroy, Vincent and Cabon, Yohann and Revaud, Jerome},
  booktitle={European Conference on Computer Vision (ECCV)},
  year={2024}
}

@article{shap-e,
  title={Shap-E: Generating conditional 3D implicit functions},
  author={Jun, Heewoo and Nichol, Alex},
  journal={arXiv preprint arXiv:2305.02463},
  year={2023}
}

@article{instantmesh,
  title={InstantMesh: Efficient 3D mesh generation from a single image with sparse-view large reconstruction models},
  author={Xu, Jiale and Cheng, Weihao and Gao, Yiming and Wang, Xintao and Gao, Shenghua and Shan, Ying},
  journal={arXiv preprint arXiv:2404.07191},
  year={2024}
}

@inproceedings{lgm,
  title={LGM: Large multi-view gaussian model for high-resolution 3D content creation},
  author={Tang, Jiaxiang and Chen, Zhaoxi and Chen, Xiaokang and Wang, Tengfei and Zeng, Gang and Liu, Ziwei},
  booktitle={European Conference on Computer Vision (ECCV)},
  year={2024}
}

@inproceedings{3dtopia-xl,
  title={3DTopia-XL: Scaling high-quality 3D asset generation via primitive diffusion},
  author={Chen, Zhaoxi and Tang, Jiaxiang and Dong, Yuhao and Cao, Ziang and Hong, Fangzhou and Lan, Yushi and Wang, Tengfei and Xie, Haozhe and Wu, Tong and Saito, Shunsuke and Pan, Liang and Lin, Dahua and Liu, Ziwei},
  booktitle={IEEE/CVF Conference on Computer Vision and Pattern Recognition (CVPR)},
  year={2025}
}

@inproceedings{trellis,
  title={Structured 3D latents for scalable and versatile 3D generation},
  author={Xiang, Jianfeng and Lv, Zelong and Xu, Sicheng and Deng, Yu and Wang, Ruicheng and Zhang, Bowen and Chen, Dong and Tong, Xin and Yang, Jiaolong},
  booktitle={IEEE/CVF Conference on Computer Vision and Pattern Recognition (CVPR)},
  year={2025}
}

@inproceedings{xcube,
  title={XCube: Large-scale 3D generative modeling using sparse voxel hierarchies},
  author={Ren, Xuanchi and Huang, Jiahui and Zeng, Xiaohui and Museth, Ken and Fidler, Sanja and Williams, Francis},
  booktitle={IEEE/CVF Conference on Computer Vision and Pattern Recognition (CVPR)},
  year={2024}
}

@inproceedings{neat,
  title={NeAT: Learning neural implicit surfaces with arbitrary topologies from multi-view images},
  author={Meng, Xiaoxu and Chen, Weikai and Yang, Bo},
  booktitle={IEEE/CVF Conference on Computer Vision and Pattern Recognition (CVPR)},
  year={2023}
}

@inproceedings{3daffordancenet,
  title={3d affordancenet: A benchmark for visual object affordance understanding},
  author={Deng, Shengheng and Xu, Xun and Wu, Chaozheng and Chen, Ke and Jia, Kui},
  booktitle={IEEE/CVF Conference on Computer Vision and Pattern Recognition (CVPR)},
  year={2021}
}

@inproceedings{where2act,
  title={Where2Act: From pixels to actions for articulated 3D objects},
  author={Mo, Kaichun and Guibas, Leonidas J and Mukadam, Mustafa and Gupta, Abhinav and Tulsiani, Shubham},
  booktitle={IEEE/CVF International Conference on Computer Vision (ICCV)},
  year={2021}
}

@inproceedings{adaafford,
  title={AdaAfford: Learning to adapt manipulation affordance for 3D articulated objects via few-shot interactions},
  author={Wang, Yian and Wu, Ruihai and Mo, Kaichun and Ke, Jiaqi and Fan, Qingnan and Guibas, Leonidas J and Dong, Hao},
  booktitle={European Conference on Computer Vision (ECCV)},
  year={2022}
}

@inproceedings{lmaffordance3d,
  title={Grounding 3D object affordance with language instructions, visual observations and interactions},
  author={Zhu, He and Kong, Quyu and Xu, Kechun and Xia, Xunlong and Deng, Bing and Ye, Jieping and Xiong, Rong and Wang, Yue},
  booktitle={IEEE/CVF Conference on Computer Vision and Pattern Recognition (CVPR)},
  year={2025}
}

@inproceedings{openad,
  title={Open-vocabulary affordance detection in 3d point clouds},
  author={Nguyen, Toan and Vu, Minh Nhat and Vuong, An and Nguyen, Dzung and Vo, Thieu and Le, Ngan and Nguyen, Anh},
  booktitle={IEEE/RSJ International Conference on Intelligent Robots and Systems (IROS)},
  year={2023}
}

@inproceedings{pointrefer,
  title={LASO: Language-guided Affordance Segmentation on 3D Object},
  author={Li, Yicong and Zhao, Na and Xiao, Junbin and Feng, Chun and Wang, Xiang and Chua, Tat-seng},
  booktitle={IEEE/CVF Conference on Computer Vision and Pattern Recognition (CVPR)},
  year={2024}
}

@inproceedings{affordancellm,
  title={AffordanceLLM: Grounding affordance from vision language models},
  author={Qian, Shengyi and Chen, Weifeng and Bai, Min and Zhou, Xiong and Tu, Zhuowen and Li, Li Erran},
  booktitle={IEEE/CVF Conference on Computer Vision and Pattern Recognition Workshops (CVPRW)},
  year={2024}
}

@article{affogato,
  title={Affogato: Learning Open-Vocabulary Affordance Grounding with Automated Data Generation at Scale},
  author={Lee, Junha and Park, Eunha and Park, Chunghyun and Kang, Dahyun and Cho, Minsu},
  journal={arXiv preprint arXiv:2506.12009},
  year={2025}
}

@article{dinov2,
  title={DINOv2: Learning Robust Visual Features without Supervision},
  author={Oquab, Maxime and Darcet, Timoth{\'e}e and Moutakanni, Th{\'e}o and Vo, Huy V and Szafraniec, Marc and Khalidov, Vasil and Fernandez, Pierre and Haziza, Daniel and Massa, Francisco and El-Nouby, Alaaeldin and Assran, Mahmoud and Ballas, Nicolas and Galuba, Wojciech and Howes, Russell and Huang, Po-Yao and Li, Shang-Wen and Misra, Ishan and Rabbat, Michael and Sharma, Vasu and Synnaeve, Gabriel and Xu, Hu and Jegou, Herv{\'e} and Mairal, Julien and Labatut, Patrick and Joulin, Armand and Bojanowski, Piotr},
  journal={Transactions on Machine Learning Research (TMLR)},
  year={2024}
}

@inproceedings{clip,
  title={Learning transferable visual models from natural language supervision},
  author={Radford, Alec and Kim, Jong Wook and Hallacy, Chris and Ramesh, Aditya and Goh, Gabriel and Agarwal, Sandhini and Sastry, Girish and Askell, Amanda and Mishkin, Pamela and Clark, Jack and Krueger, Gretchen and Sutskever, Ilya},
  booktitle={International Conference on Machine Learning (ICML)},
  year={2021}
}

@inproceedings{rectified_flow,
  title={Flow Straight and Fast: Learning to Generate and Transfer Data with Rectified Flow},
  author={Liu, Xingchao and Gong, Chengyue and Liu, Qiang},
  booktitle={International Conference on Learning Representations (ICLR)},
  year={2023}
}

@inproceedings{flow_matching,
  title={Flow Matching for Generative Modeling},
  author={Lipman, Yaron and Chen, Ricky T. Q. and Ben-Hamu, Heli and Nickel, Maximilian and Le, Matthew},
  booktitle={International Conference on Learning Representations (ICLR)},
  year={2023}
}

@inproceedings{objaversexl,
  title={Objaverse-xl: A universe of 10m+ 3d objects},
  author={Deitke, Matt and Liu, Ruoshi and Wallingford, Matthew and Ngo, Huong and Michel, Oscar and Kusupati, Aditya and Fan, Alan and Laforte, Christian and Voleti, Vikram and Gadre, Samir Yitzhak and VanderBilt, Eli and Kembhavi, Aniruddha and Vondrick, Carl and Gkioxari, Georgia and Ehsani, Kiana and Schmidt, Ludwig and Farhadi, Ali},
  booktitle={Advances in Neural Information Processing Systems (NeurIPS)},
  volume={36},
  year={2023}
}

@article{3dfuture,
  title={3d-future: 3d furniture shape with texture},
  author={Fu, Huan and Jia, Rongfei and Gao, Lin and Gong, Mingming and Zhao, Binqiang and Maybank, Steve and Tao, Dacheng},
  journal={International Journal of Computer Vision (IJCV)},
  volume={129},
  number={12},
  year={2021},
  publisher={Springer}
}

@inproceedings{hssd,
  title={Habitat synthetic scenes dataset (hssd-200): An analysis of 3d scene scale and realism tradeoffs for objectgoal navigation},
  author={Khanna, Mukul and Mao, Yongsen and Jiang, Hanxiao and Haresh, Sanjay and Shacklett, Brennan and Batra, Dhruv and Clegg, Alexander and Undersander, Eric and Chang, Angel X and Savva, Manolis},
  booktitle={IEEE/CVF Conference on Computer Vision and Pattern Recognition (CVPR)},
  year={2024}
}

@inproceedings{abo,
  title={Abo: Dataset and benchmarks for real-world 3d object understanding},
  author={Collins, Jasmine and Goel, Shubham and Deng, Kenan and Luthra, Achleshwar and Xu, Leon and Gundogdu, Erhan and Zhang, Xi and Vicente, Tomas F Yago and Dideriksen, Thomas and Arora, Himanshu and Guillaumin, Matthieu and Malik, Jitendra},
  booktitle={IEEE/CVF Conference on Computer Vision and Pattern Recognition (CVPR)},
  year={2022}
}

@inproceedings{cfg,
  title={Classifier-Free Diffusion Guidance},
  author={Ho, Jonathan and Salimans, Tim},
  booktitle={NeurIPS 2021 Workshop on Deep Generative Models and Downstream Applications},
  year={2021}
}

@inproceedings{adamw,
  title={Decoupled Weight Decay Regularization},
  author={Loshchilov, Ilya and Hutter, Frank},
  booktitle={International Conference on Learning Representations (ICLR)},
  year={2019}
}

@inproceedings{mcc,
  title={Multiview compressive coding for 3D reconstruction},
  author={Wu, Chao-Yuan and Johnson, Justin and Malik, Jitendra and Feichtenhofer, Christoph and Gkioxari, Georgia},
  booktitle={IEEE/CVF Conference on Computer Vision and Pattern Recognition (CVPR)},
  year={2023}
}

@inproceedings{dice,
  title={V-net: Fully convolutional neural networks for volumetric medical image segmentation},
  author={Milletari, Fausto and Navab, Nassir and Ahmadi, Seyed-Ahmad},
  booktitle={International Conference on 3D Vision (3DV)},
  year={2016}
}

@inproceedings{mvsnet,
title={Mvsnet: Depth inference for unstructured multi-view stereo},
author={Yao, Yao and Luo, Zixin and Li, Shiwei and Fang, Tian and Quan, Long},
booktitle={European Conference on Computer Vision (ECCV)},
year={2018}
}

@inproceedings{neuralrecon,
title={Neuralrecon: Real-time coherent 3d reconstruction from monocular video},
author={Sun, Jiaming and Xie, Yiming and Chen, Linghao and Zhou, Xiaowei and Bao, Hujun},
booktitle={IEEE/CVF Conference on Computer Vision and Pattern Recognition (CVPR)},
year={2021}
}

@inproceedings{zero1to3,
title={Zero-1-to-3: Zero-shot one image to 3d object},
author={Liu, Ruoshi and Wu, Rundi and Van Hoorick, Basile and Tokmakov, Pavel and Zakharov, Sergey and Vondrick, Carl},
booktitle={IEEE/CVF International Conference on Computer Vision (ICCV)},
year={2023}
}

@inproceedings{mvdream,
title={MVDream: Multi-view diffusion for 3D generation},
author={Shi, Yichun and Wang, Peng and Ye, Jianglong and Mai, Long and Li, Kejie and Yang, Xiao},
booktitle={International Conference on Learning Representations (ICLR)},
year={2024}
}

@inproceedings{syncdreamer,
title={SyncDreamer: Generating Multiview-consistent Images from a Single-view Image},
author={Liu, Yuan and Lin, Cheng and Zeng, Zijiao and Long, Xiaoxiao and Liu, Lingjie and Komura, Taku and Wang, Wenping},
booktitle={International Conference on Learning Representations (ICLR)},
year={2024}
}

@inproceedings{one-2-3-45,
title={One-2-3-45: Any single image to 3d mesh in 45 seconds without per-shape optimization},
author={Liu, Minghua and Xu, Chao and Jin, Haian and Chen, Linghao and Varma T, Mukund and Xu, Zexiang and Su, Hao},
booktitle={Advances in Neural Information Processing Systems (NeurIPS)},
volume={36},
year={2023}
}

@inproceedings{octmae,
  title={Zero-shot multi-object scene completion},
  author={Iwase, Shun and Liu, Katherine and Guizilini, Vitor and Gaidon, Adrien and Kitani, Kris and Ambru{\c{s}}, Rare{\c{s}} and Zakharov, Sergey},
  booktitle={European Conference on Computer Vision (ECCV)},
  year={2024}
}

@inproceedings{dreamcomposer,
  title={Dreamcomposer: Controllable 3d object generation via multi-view conditions},
  author={Yang, Yunhan and Huang, Yukun and Wu, Xiaoyang and Guo, Yuan-Chen and Zhang, Song-Hai and Zhao, Hengshuang and He, Tong and Liu, Xihui},
  booktitle={IEEE/CVF Conference on Computer Vision and Pattern Recognition (CVPR)},
  year={2024}
}

@inproceedings{ovafields,
  title={OVA-Fields: Weakly Supervised Open-Vocabulary Affordance Fields for Robot Operational Part Detection},
  author={Su, Heng and Xie, Mengying and Cao, Nieqing and Ding, Yan and Shao, Beichen and Long, Xianlei and Gu, Fuqiang and Chen, Chao},
  booktitle={IEEE/CVF International Conference on Computer Vision (ICCV)},
  year={2025}
}

@inproceedings{transmvsnet,
  title={Transmvsnet: Global context-aware multi-view stereo network with transformers},
  author={Ding, Yikang and Yuan, Wentao and Zhu, Qingtian and Zhang, Haotian and Liu, Xiangyue and Wang, Yuanjiang and Liu, Xiao},
  booktitle={IEEE/CVF Conference on Computer Vision and Pattern Recognition (CVPR)},
  year={2022}
}

@article{bundlefusion,
  title={Bundlefusion: Real-time globally consistent 3d reconstruction using on-the-fly surface reintegration},
  author={Dai, Angela and Nie{\ss}ner, Matthias and Zollh{\"o}fer, Michael and Izadi, Shahram and Theobalt, Christian},
  journal={ACM Transactions on Graphics (ToG)},
  volume={36},
  number={4},
  year={2017},
  publisher={ACM New York, NY, USA}
}

@inproceedings{toys4k,
  title={Using shape to categorize: Low-shot learning with an explicit shape bias},
  author={Stojanov, Stefan and Thai, Anh and Rehg, James M},
  booktitle={IEEE/CVF Conference on Computer Vision and Pattern Recognition (CVPR)},
  year={2021}
}

@article{sam3d,
  title={SAM 3D: 3Dfy Anything in Images},
  author={{SAM 3D Team} and Chen, Xingyu and Chu, Fu-Jen and Gleize, Pierre and Liang, Kevin J and Sax, Alexander and Tang, Hao and Wang, Weiyao and Guo, Michelle and Hardin, Thibaut and Li, Xiang and Lin, Aohan and Liu, Jiawei and Ma, Ziqi and Sagar, Anushka and Song, Bowen and Wang, Xiaodong and Yang, Jianing and Zhang, Bowen and Doll{\'a}r, Piotr and Gkioxari, Georgia and Feiszli, Matt and Malik, Jitendra},
  journal={arXiv preprint arXiv:2511.16624},
  year={2025}
}

@inproceedings{fixmatch,
  title={FixMatch: Simplifying semi-supervised learning with consistency and confidence},
  author={Sohn, Kihyuk and Berthelot, David and Li, Chun-Liang and Zhang, Zizhao and Carlini, Nicholas and Cubuk, Ekin D. and Kurakin, Alex and Zhang, Han and Raffel, Colin},
  booktitle={Advances in Neural Information Processing Systems (NeurIPS)},
  year={2020}
}

@inproceedings{flexmatch,
  title={FlexMatch: Boosting semi-supervised learning with curriculum pseudo labeling},
  author={Zhang, Bowen and Wang, Yidong and Hou, Wenxin and Wu, Hao and Wang, Jindong and Okumura, Manabu and Shinozaki, Takahiro},
  booktitle={Advances in Neural Information Processing Systems (NeurIPS)},
  year={2021}
}

@book{gibson2014ecological,
  title={The ecological approach to visual perception: classic edition},
  author={Gibson, James J},
  year={2014},
  publisher={Psychology press}
}
